\documentclass[11pt]{article}
\usepackage[margin=1in]{geometry}
\usepackage{natbib}
\usepackage{authblk}

\usepackage[utf8]{inputenc}
\usepackage[T1]{fontenc}
\usepackage{microtype}
\usepackage{hyperref}
\usepackage{url}
\usepackage{xcolor}

\usepackage{amsmath, amssymb, amsfonts}
\usepackage{amsthm}
\usepackage{mathtools}
\usepackage{nicefrac}
\usepackage{empheq}

\usepackage{graphicx}
\usepackage{subcaption}
\usepackage{caption}
\usepackage{booktabs}
\usepackage{float}
\usepackage{wrapfig}

\usepackage{algorithm}
\usepackage{algpseudocode}

\usepackage[capitalize,noabbrev]{cleveref}

\theoremstyle{plain}
\newtheorem{theorem}{Theorem}[section]
\newtheorem{proposition}[theorem]{Proposition}

\theoremstyle{definition}

\theoremstyle{remark}

\title{Discrete MeanFlow: One-Step Generation \\via Conditional Transition Kernels}

\author{Fairoz Nower Khan}
\author{Nabuat Zaman Nahim}
\author{Md Sajid Ahmed}
\author{Ruiquan Huang}
\author{Peizhong~Ju}
\affil{Department of Computer Science, University of Kentucky\\ 
\mbox{Lexington, Kentucky 40506, USA}}
\affil[ ]{fairoz.khan@uky.edu, nabuat.nahim@uky.edu, sajid.ahmed@uky.edu, ruiquanhuang@uky.edu, peizhong.ju@uky.edu}
\date{}

\begin{document}
\maketitle
\pagestyle{plain}

\begin{abstract}
MeanFlow enables one-step generation in continuous spaces by learning an average velocity
over a time interval rather than the instantaneous velocity field of flow matching.
However, discrete state spaces lack smooth trajectories or spatial derivatives,
so the continuous formulation does not directly apply.
We introduce Discrete MeanFlow, a theoretical finite-state formulation that replaces point motion with probability-mass transport.
Our key object is the conditional transition kernel
of a continuous-time Markov chain (CTMC), from which we define a mean discrete rate that measures average transition-probability change over a finite interval.
We prove a Discrete MeanFlow identity that relates this finite-interval rate
to the instantaneous CTMC generator at the endpoint, using the Kolmogorov forward equation in place of the spatial chain rule in continuous MeanFlow. We parameterize the transition kernel directly using a
boundary-by-construction design that guarantees valid probability outputs and exact
boundary conditions without auxiliary losses. Because the learned kernel defines a probability distribution, generation requires only a single forward pass followed by direct sampling, eliminating the need for iterative denoising, ODE integration, or multi-step refinement. We validate the identity on exact finite-state Markov chains, demonstrating that the learned kernel recovers the analytical ground truth with high precision. We also investigate the training objective and the limits of factorization on synthetic sequence tasks across varying alphabet sizes and sequence lengths.
\end{abstract}

\section{Introduction}
Generative models based on flow matching~\citep{lipman2022flow, liu2022rectified} and diffusion~\citep{ho2020denoising, song2021score} achieve strong results in images, audio, and molecular design, but require many ODE or SDE integration steps at inference. MeanFlow~\citep{geng2025mean} reduces this cost in continuous spaces by learning an average velocity over a time interval. An identity relating this average to the instantaneous velocity at the endpoint enables one-step generation.


However, MeanFlow does not directly extend to discrete state spaces because tokens lack smooth trajectories and spatial derivatives with respect to the state. Many established discrete diffusion and flow-matching models~\citep{austin2021structured, sahoo2024simple, lou2024discrete, campbell2024generative, gat2024discrete} learn local transition or jump-rate information and typically generate samples through multiple iterative transitions. Recent work has begun to explore one-step and few-step generation in discrete spaces, but developing principled and efficient alternatives to iterative CTMC simulation remains an open problem. These challenges are also relevant in reinforcement learning, where a recent
survey identifies extending flow-matching policies to discrete or hybrid action
spaces and improving inference efficiency as important open directions
\citep{nahim2026a}.

The core idea is to generalize the motion of probability mass rather than the motion of a state. A CTMC transition kernel describes how conditional probability mass redistributes over a finite interval. From this kernel, we define a \emph{mean discrete rate} and derive a \emph{Discrete MeanFlow identity} relating this finite-interval quantity to the endpoint CTMC generator through the Kolmogorov forward equation.

We parameterize the transition kernel by mixing a delta at the current state with a neural network output using a coefficient that vanishes when $t=r$. This \emph{boundary-by-construction} form guarantees valid probabilities and the exact boundary condition $K_{r,r}(y,x)=\delta_{y,x}$ without a penalty loss.

Once a full-interval kernel is learned, endpoint sampling requires one network evaluation and direct sampling from the predicted kernel, without iterative CTMC simulation. For high-dimensional sequences, however, representing the full joint kernel is intractable, so practical parameterizations introduce structural approximations whose limitations we analyze explicitly.

\begin{figure}[t]
    \centering
    \includegraphics[width=\linewidth]{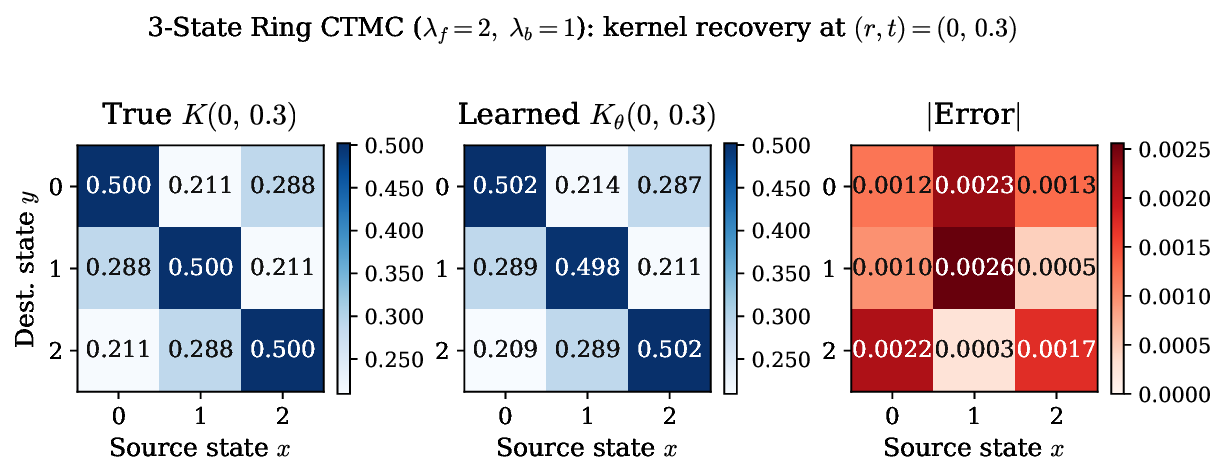}
    \caption{%
    \textbf{Exact kernel recovery on a 3-state ring CTMC.}
    The learned kernel $K_\theta$ closely matches the analytical kernel $K$, with
    maximum absolute error below $3\times 10^{-3}$ and accurate recovery of the
    ring's asymmetric transition structure.
    }
    \label{fig:intro_heatmap}
\end{figure}

We first use exact finite-state CTMCs as proof-of-concept tests of the identity. Figure~\ref{fig:intro_heatmap} shows one such exact-kernel recovery result on a 3-state ring CTMC. We then use controlled sequence models to analyze two obstacles to strict one-step generation. Under masking, the naive kernel-residual target has an endpoint singularity, while Rao-Blackwellization removes it. For correlated sequences, tokenwise factorization imposes a total-correlation floor that no training objective can overcome, while block factorization lowers this floor by representing local dependencies.

Our main contributions are:

\begin{enumerate}
    \item We formulate \emph{Discrete MeanFlow} for finite state spaces and derive an identity connecting a finite-interval mean discrete rate to the instantaneous CTMC generator through the Kolmogorov forward equation.

    \item We analyze the kernel-residual objective under masking, show that its naive finite-sample form has an endpoint singularity, and derive a Rao-Blackwellized target that removes it.

    \item We characterize the structural limit of tokenwise one-step generation through a total-correlation lower bound and show how block factorization lowers this floor by representing local dependencies.
\end{enumerate}

\section{Related Work}

\paragraph{Few step generation.}
Diffusion~\citep{sohl2015deep, ho2020denoising, song2019generative, song2021score} and flow-matching~\citep{lipman2022flow, liu2022rectified, albergo2023building} models are effective continuous-space generators but typically require iterative sampling. Consistency Models~\citep{song2023consistency, song2024improved}, Shortcut Models~\citep{frans2025one}, and Inductive Moment Matching~\citep{zhou2025inductive} enable one or few-step continuous generation using consistency or finite-interval training objectives.
MeanFlow~\citep{geng2025mean} instead relates average and instantaneous velocities through a finite-interval identity, with subsequent extensions and analyses~\citep{sheng2026mp1,wang2026one,chen2025one,li2025meanaudio,geng2025improved,zhang2025alphaflow}. These formulations assume continuous state trajectories and therefore do not directly transfer to finite discrete state spaces.

\paragraph{Discrete generative models.}
Discrete diffusion models such as D3PM~\citep{austin2021structured}, SEDD~\citep{lou2024discrete}, and MDLM~\citep{sahoo2024simple}, and Discrete Flow Matching~\citep{gat2024discrete,campbell2024generative}, model local reverse transitions or CTMC rates and typically rely on iterative generation. CTMC-based discrete flow models have also been used as generative policies 
for discrete-action reinforcement learning, including Q-weighted offline 
policy learning \citep{khan2026flow} and offline-to-online fine-tuning with 
path-space regularization \citep{khan2026discrete}. Recent work accelerates discrete generation through Gaussian-to-discrete consistency distillation~\citep{sahoo2025diffusion}, coupling rectification that reduces factorization error~\citep{yoo2026redi}, direct latent--data coupling models~\citep{peng2026coupling}, and categorical or discrete flow maps~\citep{roos2026categorical,lee2026flow,potaptchik2026discrete}. Our focus is complementary. We derive a finite-state MeanFlow identity directly for CTMC transition kernels and use it to analyze the statistical and structural obstacles that arise when learning full-interval kernels for one-step generation. In particular, our contribution is not that a denoiser can be sampled once from a fully corrupted input, but the finite-interval CTMC identity and the resulting analysis of estimator variance and kernel factorization.

\section{Background}
\label{sec:background}

\begin{figure}[t]
    \centering
    \includegraphics[width=\linewidth]{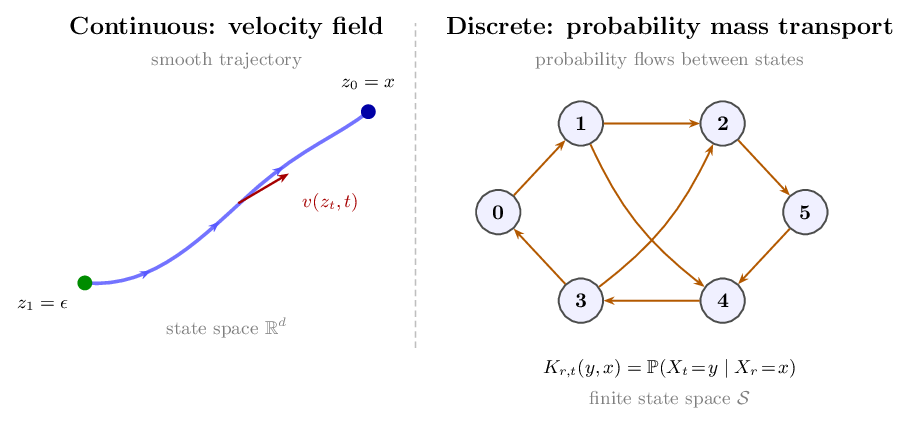}
    \caption{%
    \textbf{Continuous vs.\ discrete generative dynamics.}
    Continuous flow matching follows smooth trajectories, whereas discrete dynamics
    redistribute probability mass over finite states using a CTMC transition kernel
    $K_{r,t}(y,x)$.
    }
    \label{fig:background}
\end{figure}

This section reviews the continuous foundations our work generalizes, flow matching and MeanFlow, before introducing continuous-time Markov chains as the discrete counterpart.
Figure~\ref{fig:background} highlights the key contrast where continuous models move points along smooth velocity-driven trajectories, whereas discrete models describe probability mass flowing between states.

\subsection{Flow Matching}

Flow matching~\citep{lipman2022flow, liu2022rectified}
learns a velocity field that transports a prior distribution to a data
distribution.
Given data $x \sim p_{\mathrm{data}}$ and noise
$\epsilon \sim p_{\mathrm{prior}}$, a flow path is constructed as
$z_t = (1-t)x + t\epsilon$, with conditional velocity
$v_t = \epsilon - x$.
Because a given $z_t$ can arise from many different $(x, \epsilon)$ pairs,
flow matching trains a network $v_\theta$ to fit the \emph{marginal} velocity
field $v(z_t, t) = \mathbb{E}[v_t \mid z_t]$.
Samples are generated by solving the ODE $\tfrac{d}{dt}z_t = v(z_t, t)$
from $z_1 = \epsilon$ to $z_0 = x$.

Even with straight conditional flows, the marginal velocity field can induce curved trajectories~\citep{lipman2022flow}. Coarse discretization then yields inaccurate samples, so standard flow matching requires many ODE solver steps at generation time.

\subsection{MeanFlow}

MeanFlow~\citep{geng2025mean} replaces the instantaneous velocity with an
\emph{average velocity} over a time interval $[r, t]$:
\[
u(z_t,r,t)
=
\frac{1}{t-r}\int_r^t v(z_\tau,\tau)\,d\tau.
\]
This field describes the average direction and magnitude of the flow between
two time steps.
Unlike the instantaneous velocity, the average velocity directly encodes
displacement: $z_r = z_t - (t-r)\,u(z_t, r, t)$.
A single evaluation of $u(\epsilon, 0, 1)$ maps noise to data in
one step, without integrating the ODE. The key theoretical result of MeanFlow is an identity relating the average
and instantaneous velocities.
Differentiating the definition of $u$ with respect to $t$ and applying the
product rule yields:
\[
u(z_t,r,t)
=
v(z_t,t)
-
(t-r)\frac{d}{dt}u(z_t,r,t).
\]
The MeanFlow identity states that the average velocity equals the instantaneous velocity at the endpoint minus a correction proportional to how the average velocity changes with time.
The total derivative $\tfrac{d}{dt}u$ expands via the chain rule into
$v(z_t,t)\partial_z u + \partial_t u$, involving a spatial Jacobian-vector
product.
A neural network $u_\theta$ is trained to satisfy this identity using the instantaneous velocity as the ground-truth signal, without computing the integral during training.

\subsection{Continuous-Time Markov Chains}

A continuous-time Markov chain
(CTMC)~\citep{norris1998markov} on a finite state space $\mathcal{S}$ is
specified by a time-dependent generator $u_t(y,x)$, which gives the
instantaneous rate at which the process jumps from state $x$ to state $y$.
The generator satisfies $u_t(y,x) \ge 0$ for $y \neq x$ and
$\sum_{y \in \mathcal{S}} u_t(y,x) = 0$, so that each column sums to zero
and probability is conserved.

\paragraph{Transition
kernel} The finite-time behavior of a CTMC is captured by the \emph{transition
kernel}
\[
K_{r,t}(y,x)
=
\mathbb{P}(X_t=y\mid X_r=x).
\]
which gives the probability of being in state $y$ at time $t$ given that
the process was in state $x$ at time $r$.
For each starting state $x$, the column $K_{r,t}(\cdot, x)$ is a valid
probability distribution.
When no time elapses, the process stays put:
$K_{r,r}(y,x) = \delta_{y,x}$.

The generator and the kernel are linked by the Kolmogorov forward equation~\citep{lipman2024flow}:
\[
\partial_t K_{r,t}(y,x)
=
\sum_{z\in\mathcal{S}}
u_t(y,z)K_{r,t}(z,x).
\]
This equation describes how conditional probabilities evolve, with each state $z$ contributing its current probability mass weighted by the instantaneous rate from $z$ to $y$. The Kolmogorov forward equation therefore links the kernel derivative to the instantaneous generator. Thus, CTMC dynamics describe the evolution of probability mass rather than the motion of a state. Our discrete formulation builds on this relation by learning finite-interval transition kernels rather than simulating local transitions.

\section{Discrete MeanFlow}
\label{sec:discrete_meanflow}

Extending one-step generation to discrete spaces is not straightforward.
Continuous flow matching and MeanFlow rely on three ingredients that do not exist in discrete settings: smooth trajectories through state space,
velocity fields that describe motion along those trajectories, and spatial
derivatives that appear in the chain rule when differentiating along a flow.
A token cannot move smoothly from one value to another, so none of these
tools transfer directly.

The central challenge lies in determining the appropriate discrete analogue. One strategy is to relax discrete states into continuous representations and apply continuous-space methods. Our goal instead is a finite-state formulation that preserves the underlying CTMC structure. We therefore argue that the correct object to work with is not the motion of a state but the motion of \emph{probability mass} across states. This shift in perspective, from tracking the trajectory of a point to tracking the evolution of a distribution, is what makes a principled discrete theory possible.

Concretely, a continuous-time Markov chain is fully characterized by its
transition kernel, which records how probability redistributes over a
finite state space between any two times. We show that this kernel admits a mean discrete rate that plays the role of MeanFlow's average velocity. We further prove that this rate satisfies a closed-form identity derived from the Kolmogorov forward equation, replacing the chain rule and spatial Jacobian used in the continuous setting. Finally, we show that learning the kernel directly enables one-step discrete generation. Each step of this development requires care as the Kolmogorov forward equation is not a spatial chain rule, and its role in the identity is structurally different from the role of ODE dynamics in the continuous MeanFlow identity.

\subsection{Mean Discrete Rate}
\label{sec:mean-rate}
The starting point is the transition kernel $K_{r,t}(y,x)$, which records the
probability of arriving at state $y$ at time $t$ given that the process started
in state $x$ at time $r$.
At the boundary $t = r$, the process has not moved, so $K_{r,r}(y,x) = \delta_{y,x}$.
The total change in conditional probability over the interval $[r, t]$ is therefore
$K_{r,t}(y,x) - \delta_{y,x}$.
Dividing by the interval length gives the average rate at which probability mass
has moved toward state $y$:
\begin{equation}
    \bar{u}(y, x, r, t)
    \;:=\;
    \frac{K_{r,t}(y,x) - \delta_{y,x}}{t - r}.
    \label{eq:mean_rate}
\end{equation}
This is the \emph{mean discrete rate} which replaces the average velocity in continuous MeanFlow, and it is a
finite-interval description of dynamics, not an instantaneous one
(Figure~\ref{fig:method}). The mean discrete rate inherits structure from the kernel it is derived from.
Its off-diagonal entries are non-negative because $K_{r,t}(y,x)\geq0$, its diagonal entry is non-positive because $K_{r,t}(x,x)\leq1$, and its column sums to zero because probability is conserved.
Crucially, the kernel can be recovered by inverting the definition:
$K_{r,t}(y,x) = \delta_{y,x} + (t-r)\,\bar{u}(y,x,r,t)$,
and for the true mean rate this reconstruction recovers the valid transition kernel exactly.

\subsection{The Discrete MeanFlow Identity}
\label{sec:discrete-meanflow-identity}


The central result of this work is an identity that connects the mean discrete rate
$\bar{u}(y, x, r, t)$ to the instantaneous generator $u_t$ of the CTMC. It provides the foundation
for the training objective developed in Section~\ref{sec:training}.

\textbf{Starting point: the kernel form.}
By Eq.~\eqref{eq:mean_rate}, the transition kernel admits the decomposition
\begin{equation}
K_{r,t}(y, x) \;=\; \delta_{y,x} \;+\; (t - r)\,\bar{u}(y, x, r, t),
\label{eq:kernel-form}
\end{equation}

\textbf{Differentiating in the endpoint time.}
We differentiate both sides of \eqref{eq:kernel-form} with respect to $t$, holding
$r$ and $x$ fixed. The left-hand side yields $\partial_t K_{r,t}(y, x)$. On the
right-hand side, $\delta_{y,x}$ is constant in $t$, and the product rule applied
to $(t-r)\,\bar{u}(y, x, r, t)$ gives
\begin{equation}
\partial_t K_{r,t}(y, x)
\;=\; \bar{u}(y, x, r, t) \;+\; (t - r)\,\partial_t \bar{u}(y, x, r, t).
\label{eq:derivative-kernel-form}
\end{equation}

The transition kernel of the CTMC also satisfies the Kolmogorov forward equation
\begin{equation}
\partial_t K_{r,t}(y, x) \;=\; \sum_{z \in \mathcal{S}} u_t(y, z)\, K_{r,t}(z, x),
\quad K_{r,r}(y, x) = \delta_{y,x},
\label{eq:forward-equation}
\end{equation}
This describes how probability flows into state $y$ where each state $z$ contributes its current mass $K_{r,t}(z,x)$, weighted by the instantaneous transfer rate $u_t(y,z)$ from $z$ to $y$

Equating \eqref{eq:derivative-kernel-form} and \eqref{eq:forward-equation},
\begin{equation}
\bar{u}(y, x, r, t) \;+\; (t - r)\,\partial_t \bar{u}(y, x, r, t)
\;=\; \sum_{z \in \mathcal{S}} u_t(y, z)\, K_{r,t}(z, x).
\label{eq:combined}
\end{equation}

\textbf{Recognizing the conditional expectation.}
Since $K_{r,t}(z, x) = \mathbb{P}(X_t = z \mid X_r = x)$, the right-hand side of
\eqref{eq:combined} is precisely the conditional expectation of $u_t(y, X_t)$
given $X_r = x$:
\begin{equation}
\sum_{z \in \mathcal{S}} u_t(y, z)\, K_{r,t}(z, x)
\;=\; \mathbb{E}\!\left[ u_t(y, X_t) \,\big|\, X_r = x \right].
\label{eq:expectation-form}
\end{equation}
Substituting \eqref{eq:expectation-form} into \eqref{eq:combined} and rearranging
yields the main result.

\begin{theorem}[Discrete MeanFlow Identity]
\label{thm:discrete-meanflow}
Let $(X_t)_{t \in [0,1]}$ be a finite-state CTMC with bounded generator $u_t$
that is continuous in $t$. Then for every $x, y \in \mathcal{S}$ and every
$0 \leq r < t \leq 1$,
\begin{empheq}[box=\fbox]{equation}
\begin{split}
    \underbrace{\bar{u}(y,x,r,t)}_{\text{mean rate}}
    &=\;
    \underbrace{\mathbb{E}\!\left[u_t(y, X_t) \mid X_r = x\right]}_{\text{instantaneous (expected)}}
    \\[-2pt]
    &\quad-\;
    \underbrace{(t-r)\,\partial_t \bar{u}(y,x,r,t)}_{\text{time derivative correction}}
\end{split}
\label{eq:dmf_identity}
\end{empheq}
\end{theorem}

Here, the mean discrete rate equals the expected instantaneous generator
minus a correction involving $\partial_t \bar{u}$. In discrete spaces, there is no moving state and no spatial derivative.
Instead, the Kolmogorov forward equation provides the link between the kernel's
time derivative and the instantaneous generator, and the correction involves
only $\partial_t \bar{u}$, a partial derivative with respect to the endpoint
time.

An equivalent integral form, derived in
Appendix~\ref{app:integral_form}, shows that $\bar{u}$ is the time-average
of instantaneous probability flow over $[r,t]$.

\paragraph{Endpoint regularity.}
The bounded-generator assumption in Theorem~\ref{thm:discrete-meanflow} is needed only on the
time interval over which the identity is applied. In particular, for
processes whose rates become singular at the endpoint $t=1$, the same
argument applies on every compact interval $[0,T]$ with $T<1$.
This distinction will be relevant for the masking process considered
below, whose reverse rates scale as $1/(1-t)$ near the clean-data
endpoint.

\subsection{Kernel Parameterization}
\label{sec:kernel_parameterization}
To learn the transition kernel, we parameterize it with a design that enforces
the boundary condition by construction.
A neural network $f_\theta(x, r, t)$ outputs logits over destination states,
which are passed through a softmax to produce a distribution
$q_\theta(y \mid x, r, t)$.
The learned kernel is then defined as

\begin{empheq}[box=\fbox]{equation}
\begin{aligned}
K_\theta(y,x,r,t)
&=
\bigl(1-\alpha(r,t)\bigr)\delta_{y,x}
\\
&\quad+
\alpha(r,t)q_\theta(y\mid x,r,t).
\end{aligned}
\label{eq:kernel_param}
\end{empheq}

where $\alpha(r,t) \in [0,1]$ is a scalar function satisfying $\alpha(r,r) = 0$.
At $t = r$, the mixing coefficient vanishes and the kernel reduces to $\delta_{y,x}$,
so the boundary condition holds exactly regardless of what $q_\theta$ predicts.
To obtain a nondegenerate finite boundary rate, we use schedules that
vanish linearly as $t\to r$. The schedules used in our experiments are
described in Appendix~\ref{app:boundary_param}.
This parameterization simultaneously guarantees valid probability
outputs and exact boundary satisfaction without an auxiliary boundary
loss.

\subsection{Training}
\label{sec:training}
The Discrete MeanFlow identity provides a natural training objective.
If the model is correct, the time derivative of the learned kernel should equal
the expected instantaneous generator.
We approximate the expectation with a single Monte Carlo sample, given
$x_r \sim p_r$ and $x_t \sim K_{r,t}(\cdot, x_r)$, the quantity
$u_t(y, x_t)$ is an unbiased estimator of
$\mathbb{E}[u_t(y, X_t) \mid X_r = x_r]$.
The training loss is the squared residual between the kernel's time derivative
and this stochastic target:
\begin{equation}
    \mathcal{L}(\theta)
    \;=\;
    \mathbb{E}\!\left[
    \sum_{y \in \mathcal{S}}
    \bigl(\partial_t K_\theta(y, x_r, r, t) - u_t(y, x_t)\bigr)^2
    \right].
    \label{eq:loss}
\end{equation}

The derivative $\partial_t K_\theta$ is obtained by automatic
differentiation with respect to the endpoint time $t$.
Training samples $X_t$ are drawn from the fixed reference kernel
$K_{r,t}(\cdot,X_r)$, so Eq.~\eqref{eq:loss} gives an unbiased stochastic
estimate of the population objective. Full training and reference-sampling
details are given in Appendix~\ref{app:training_algorithm}.

\paragraph{Data-conditional training.}
In data-driven settings, the marginal generator induced by corruption is
generally unknown. Let
$u_t^{\mathrm{cond}}(\cdot,\cdot;X^\star)$ denote the generator
conditioned on clean data $X^\star$, and let
\[
u_t^{\mathrm{marg}}(y,x)
=
\mathbb{E}\!\left[
u_t^{\mathrm{cond}}(y,x;X^\star)\mid X_t=x
\right].
\]
We compare the weighted kernel-residual objectives obtained using these
marginal and sample-wise conditional targets. The full process assumptions
and loss definitions are given in Appendix~\ref{app:conditional_kr_proof}.




\begin{proposition}[Data-conditional kernel-residual equivalence]
\label{prop:conditional_kr}
Assume the unconditional corruption process is Markov, $(r,t)$ is sampled
independently with $r<t<1$, and the relevant weighted second moments are
finite. Then
\begin{align}
\mathcal{L}^{w}_{\mathrm{cond}}(\theta)
&=
\mathcal{L}^{w}_{\mathrm{marg}}(\theta)
\nonumber\\
&\quad+
\mathbb{E}\!\left[
w(r,t)\sum_{y\in\mathcal{S}}
\operatorname{Var}\!\left(
u_t^{\mathrm{cond}}(y,X_t;X^\star)
\,\middle|\,X_t,t
\right)
\right].
\label{eq:conditional_kr_decomposition}
\end{align}
The second term is independent of $\theta$, so the two objectives have
the same population minimizers. Over unrestricted measurable predictors,
their pointwise minimizer is
$\partial_t K_{r,t}(y,x)$.
\end{proposition}

The full assumptions, loss definitions, and proof are given in
Appendix~\ref{app:conditional_kr_proof}.
The linear independent-masking process used in our sequence experiments
satisfies the required Markov assumption; see
Appendix~\ref{app:conditional_kr_proof}.

Proposition~\ref{prop:conditional_kr} removes the need to evaluate the
unknown marginal generator during data-driven training. Instead, one
samples $X^\star$, generates $(X_r,X_t)$ using the known corruption
process, and regresses on the computable target
$u_t^{\mathrm{cond}}(\cdot,X_t;X^\star)$. The price of this replacement
is the additional conditional-variance term in
Eq.~\eqref{eq:conditional_kr_decomposition}. Under masking, this
finite-sample target becomes singular near $t=1$. We next show that
conditioning on information already available during training removes
this singular component exactly.

\begin{proposition}[Rao-Blackwellized kernel-residual target]
\label{prop:rb}
For
\[
\bar U_M(y)
=
\frac{1}{M}
\sum_{j=1}^{M}
u_t^{\mathrm{cond}}(y,X_t^{(j)};X^\star),
\]
let $X_t^{(1)},\ldots,X_t^{(M)}$ be conditionally independent draws
from the corruption process given $(X_r,X^\star,r,t)$, and define
\[
\widetilde U(y)
=
\mathbb{E}\!\left[
u_t^{\mathrm{cond}}(y,X_t;X^\star)
\,\middle|\,
X_r,X^\star,r,t
\right].
\]

\textbf{(i) Same population target.}
For every $X_r=x$,
\[
\begin{aligned}
\mathbb{E}\!\left[
\widetilde U(y)\mid X_r=x,r,t
\right]
&=
\mathbb{E}\!\left[
\bar U_M(y)\mid X_r=x,r,t
\right]
\\
&=
\partial_t K_{r,t}(y,x).
\end{aligned}
\]
Thus, both targets have the same pointwise population minimizer.

\textbf{(ii) Closed form under linear masking.}
For coordinate $d$, let $m$ denote the mask state. If $X_{r,d}$ is
already revealed, then $\widetilde U_d(y)=0$. If $X_{r,d}=m$, then
\begin{equation}
\widetilde U_d(y)
=
\frac{
\mathbf{1}\{y=X^\star_d\}
-
\mathbf{1}\{y=m\}
}{1-r}.
\label{eq:rb_target}
\end{equation}

\textbf{(iii) Second moment.}
For one coordinate under linear independent masking with a fully masked
source,
\begin{equation}
\mathbb{E}\!\left[
\sum_y \bar U_M(y)^2
\,\middle|\,
r,t
\right]
=
\frac{2}{1-r}
+
\frac{2(t-r)}
{M(1-r)(1-t)}.
\label{eq:second_moment}
\end{equation}
\end{proposition}

Equation~\eqref{eq:second_moment} explains the earlier endpoint
singularity. At $M=1$, it simplifies to $2/(1-t)$, recovering the
single-sample result. Increasing $M$ reduces the singular term by
$1/M$, but no finite $M$ removes it. In the Rao-Blackwellized limit,
only $2/(1-r)$ remains, which is bounded under the enforced minimum
time gap.

More precisely, the second term in Eq.~\eqref{eq:second_moment} is the
irreducible conditional variance caused by sampling $X_t$ after
conditioning on $(X_r,X^\star,r,t)$. Therefore, any squared-residual
objective using a finite number of endpoint samples has population risk
at least this term. If the time sampler has positive density near
$t=1$, the unweighted population loss is non-integrable for every
finite $M$. Rao-Blackwellization removes this source of divergence.

We call the objective obtained by replacing the stochastic target in
Eq.~\eqref{eq:loss} with $\widetilde U$ the
Rao-Blackwellized kernel-residual objective (KR-RB).

A second way to control the singularity is to downweight samples near
the endpoint. Using $w(t)=1-t$ gives the weighted conditional
kernel-residual objective
\begin{equation}
\begin{split}
\mathcal{L}_{\mathrm{WCKR}}(\theta)
&=
\mathbb{E}\!\Biggl[
(1-t)
\sum_{y\in\mathcal{S}}
\Bigl(
\partial_t K_\theta(y,X_r,r,t)
\\[-2pt]
&\quad-\;
u_t^{\mathrm{cond}}(y,X_t;X^\star)
\Bigr)^2
\Biggr].
\end{split}
\label{eq:wckr}
\end{equation}
By Proposition~\ref{prop:conditional_kr}, this weighting preserves the
same pointwise population minimizer for every $t<1$ wherever the
relevant weighted moments are finite. Unlike Rao-Blackwellization,
however, it controls the singularity by reducing the contribution of
samples near $t=1$, where one-step generation is evaluated. We compare
these alternatives empirically in Section~\ref{sec:experiments}.


\subsection{One-Step Generation}

Given a starting state $x_0$, we evaluate
$K_\theta(\cdot,x_0,0,1)$ and sample directly from the resulting
endpoint distribution. Thus, generation requires one network evaluation
and no iterative CTMC simulation. For sequence models, the corresponding
sampling rule depends on the chosen factorization of the joint kernel.


\subsection{Sequence Factorization and Its Limitation}
\label{sec:factorized-extension}

For sequence generation over product state spaces $\mathcal{S} = \mathcal{V}^D$,
the full joint kernel $K_{r,t}(y, x)$ is intractable, with output dimension
$|\mathcal{V}|^D$. We therefore factorize the kernel across positions,
\begin{equation}
K_\theta(y, x, r, t) \;\approx\; \prod_{d=1}^{D} K_\theta^{(d)}(y_d \mid x, r, t),
\label{eq:factorized-kernel}
\end{equation}
where each factor predicts a per-position distribution over $y_d \in \mathcal{V}$
conditioned on the full input $x$. This reduces the output space from
$|\mathcal{V}|^D$ to $D \cdot |\mathcal{V}|$.

\paragraph{Exact coordinate-wise case.}
The factorization in Eq.~\eqref{eq:factorized-kernel} is exact when the
generator decomposes coordinate-wise, so that only one coordinate changes
per jump and its rate depends only on that coordinate. In this case, the
joint kernel factorizes and each coordinate satisfies its own Discrete
MeanFlow identity. Independent masking and per-position uniform or
absorbing corruption processes have this structure when conditioned on
the clean endpoint. The corresponding generator decomposition and
coordinate-wise identity are given in
Appendix~\ref{app:factorized_exact}.

However, this does not imply that the corresponding
data-marginal reverse kernel factorizes when the data distribution
contains dependencies across coordinates. In that setting, the
product-form parameterization in
Eq.~\eqref{eq:factorized-kernel} is an approximation: although each
factor may condition on the full current input, the output distribution
at a fixed input remains a product across destination coordinates.

\paragraph{Limitation of deterministic-source one-step factorization.}
The limitation of product-form transitions in few-step discrete
generation has been studied in recent work~\citep{yoo2026redi}.
For strict one-step generation from a deterministic source, the
factorization of the output distribution imposes a structural error
floor that cannot be removed by improving the training objective.

To interpolate between tokenwise and fully joint prediction, partition
the sequence into $J=D/B$ contiguous non-overlapping blocks
$\{\mathcal{B}_b\}_{b=1}^{J}$ of width $B$ and use
\[
K_\theta(y,x,r,t)
\approx
\prod_{b=1}^{J}
K_\theta^{(b)}
(y_{\mathcal{B}_b}\mid x,r,t).
\]
The case $B=1$ recovers tokenwise factorization, while $B=D$ recovers a
fully joint output. During masking training, each block head has
$(|\mathcal{V}|+1)^B$ possible outputs, so practical models use small
block sizes.

\begin{proposition}[Factorization error for one-step generation]
\label{prop:factorization_lower_bound}
Let $p$ be a distribution over
$X=(X_1,\ldots,X_D)\in\mathcal{V}^D$, let $B$ divide $D$, and let
$\{\mathcal{B}_b\}_{b=1}^{D/B}$ be contiguous blocks of width $B$.
Suppose one-step generation starts from a deterministic source and
produces
\[
q(x)
=
\prod_{b=1}^{D/B}
q_b(x_{\mathcal{B}_b}),
\]
where $q_b(x_{\mathcal{B}_b})>0$ whenever
$p_{\mathcal{B}_b}(x_{\mathcal{B}_b})>0$. Then
\begin{equation}
D_{\mathrm{KL}}(p\|q)
=
\mathrm{TC}_B(p)
+
\sum_{b=1}^{D/B}
D_{\mathrm{KL}}
\!\left(p_{\mathcal B_b}\|q_b\right).
\label{eq:factorization_kl_decomposition}
\end{equation}
where
\begin{equation}
\mathrm{TC}_B(p)
:=
\sum_{b=1}^{D/B}
H(X_{\mathcal{B}_b})
-
H(X_{1:D})
\label{eq:total_correlation}
\end{equation}
is the total correlation across blocks. Consequently,
\begin{equation}
\min_{q=\prod_b q_b}
D_{\mathrm{KL}}(p\|q)
=
\mathrm{TC}_B(p),
\label{eq:factorization_minimum}
\end{equation}
with the minimum attained by
$q_b=p_{\mathcal{B}_b}$ for every block.

If $p$ is a stationary first-order Markov chain, then
\begin{equation}
\mathrm{TC}_B(p)
=
\left(\frac{D}{B}-1\right)
I(X_{d-1};X_d).
\label{eq:markov_block_tc}
\end{equation}
\end{proposition}

Tokenwise factorization is the special case $B=1$, for which
$\mathrm{TC}_1(p)=\mathrm{TC}(p)$. For stationary first-order Markov
data, increasing $B$ therefore lowers the structural one-step error
floor by representing dependencies within each block.

The deterministic-source assumption is essential. If the source itself
is random, marginalizing a block-factorized conditional distribution
over the source produces a mixture of block-product distributions,
which can represent dependencies across blocks.

\paragraph{Blockwise Rao-Blackwellized target.}
Block-factorized models can be trained using the same
Rao-Blackwellization principle as Proposition~\ref{prop:rb}.
Consider a block containing $m$ coordinates that are masked in $X_r$.
For a feasible destination $y_{\mathcal{B}}$, let $S$ be the subset of
these coordinates that take their clean values in
$y_{\mathcal{B}}$, let $k=|S|$, and define
\[
p_{r,t}
=
\frac{t-r}{1-r}.
\]
The Rao-Blackwellized block target is
\begin{equation}
\widetilde U_{\mathcal{B}}(y_{\mathcal{B}})
=
\frac{
p_{r,t}^{\,k-1}
(1-p_{r,t})^{m-k}
\bigl(k-mp_{r,t}\bigr)
}{1-t}.
\label{eq:block_rb_target}
\end{equation}
The target is zero when $m=0$ or when
$y_{\mathcal{B}}$ is inconsistent with the revealed coordinates and
clean endpoint. A derivation is given in
Appendix~\ref{app:block_rb}.

\section{Experiments}
\label{sec:experiments}

\begin{figure*}[t]
    \centering
    \includegraphics[width=0.9\textwidth]{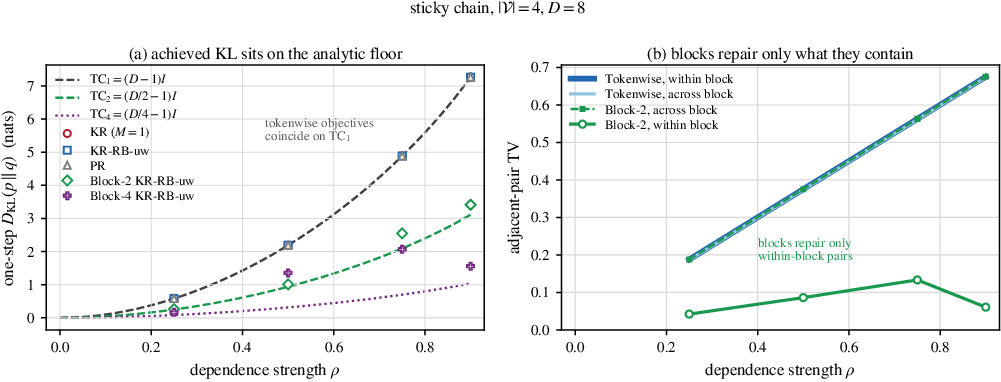}
    \caption{%
    \textbf{Block factorization lowers the structural one-step error floor.}
    \textit{Left:} tokenwise models attain $\mathrm{TC}_1$, while blocks lower
    the floor as predicted by
    Proposition~\ref{prop:factorization_lower_bound}.
    \textit{Right:} Block-2 recovers within-block but not cross-block
    dependence. Markers show three-seed means; seed variability is reported
    in Appendix~\ref{app:statistics}.
    }
    \label{fig:block_floor}
\end{figure*}

We use controlled experiments to test three predictions of the theory:
whether the Discrete MeanFlow objective recovers exact finite-state
kernels, whether Rao-Blackwellization removes the masking endpoint
singularity, and whether one-step sequence generation is limited by the
factorization of the output kernel. Full experimental details and
additional results are given in Appendix~\ref{app:experimental_details}.

\subsection{Exact Kernel Recovery}
\label{sec:exact_kernel_recovery}

We first test the identity on CTMCs whose transition kernels are known
analytically: a 2-state symmetric chain, a 3-state asymmetric ring, and a
10-state birth--death chain. Across the full $(r,t)$ grid, the maximum
entrywise kernel errors are $1.4\times10^{-4}$, $7.1\times10^{-3}$, and
$1.3\times10^{-2}$, respectively. Together with
Figure~\ref{fig:intro_heatmap}, these results provide a direct
proof-of-concept validation that the kernel-residual objective can recover
the finite-interval transition kernel. Full heatmaps and convergence
results are given in Appendix~\ref{app:stage1}.

\subsection{Rao-Blackwellizing the Kernel Residual}
\label{sec:rb_experiments}

We next study masking-based sequence generation. We compare the naive
kernel-residual target using $M\in\{1,4,16\}$ endpoint samples, its
Rao-Blackwellized limit (KR-RB), and a corrected version that removes the
$1/(1-r)^2$ weighting induced by the masking schedule (KR-RB-uw).
Posterior regression (PR) and kernel cross-entropy (KCE) provide direct
supervision baselines. Across six independent and bigram configurations,
KR-RB-uw improves one-step marginal TV over naive KR by
$3.5$--$21\times$; complete results are in
Appendix~\ref{app:corrected_sequence_results}.

\begin{figure}[t]
    \centering
    \includegraphics[width=0.9\linewidth]{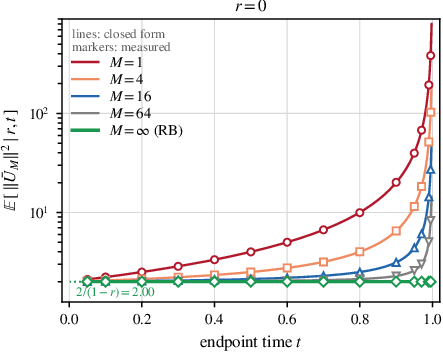}
    \caption{%
    \textbf{Rao-Blackwellization removes the endpoint singularity.}
    Closed-form second moments from Eq.~\eqref{eq:second_moment} and measured
    values at $r=0$. Each marker averages $20{,}000$ corruption samples.
    Finite $M$ diverges as $t\to1$, while RB remains bounded.
    }
    \label{fig:second_moment}
\end{figure}

Figure~\ref{fig:second_moment} directly tests
Proposition~\ref{prop:rb}. The measured second moments follow
Eq.~\eqref{eq:second_moment} to within $1.5\%$ except at four extreme
cells with $t\ge0.99$ and $M\le4$, where the Monte Carlo estimate itself
is noisy. Increasing $M$ reduces the singular term by $1/M$, whereas
Rao-Blackwellization removes it analytically. Results for
$r\in\{0,0.3,0.6\}$ are shown in
Appendix~\ref{app:second_moment_full}.

\subsection{The Factorization Floor}
\label{sec:factorization_floor}

We isolate structural error using the sticky Markov chain
$P_\rho(j\mid i)=\rho\mathbf{1}\{j=i\}+(1-\rho)/|\mathcal V|$.
Its stationary marginals are uniform, while $\rho$ controls dependence.
Because strict one-step generation begins from the same fully masked
source, a tokenwise model produces a product distribution. Across a
$42\times$ range of total correlation, all tokenwise objectives attain
the analytic floor $\mathrm{TC}(p)$ to within $0.9\%$, while generated
adjacent mutual information remains essentially zero.

Figure~\ref{fig:block_floor} shows the corresponding block result.
At $\rho=0.9$, Block-2 lowers total KL from $7.263$ to $3.413$ and
reduces within-block adjacent-pair TV from $0.675$ to $0.061$, while
across-block TV remains $0.675$. Thus blocks repair the dependencies
they represent rather than changing the training objective alone.
The same behavior holds for $|\mathcal V|=8$, $D=16$, and wider-block
results are reported in Appendix~\ref{app:block_results}.

\section{Conclusion}

We introduced Discrete MeanFlow, a finite-state formulation of MeanFlow based on CTMC transition kernels. The derived identity provides a principled objective for one-step kernel learning and enables recovery of exact finite-state transition kernels in controlled settings. Our analysis highlights both a statistical challenge and a structural limit in strict one-step sequence generation. Under masking, Rao-Blackwellization eliminates the singular finite-sample component of the kernel-residual target. For deterministic sources, tokenwise generation is subject to a total-correlation floor that no training objective can overcome. Block factorization lowers this floor by capturing dependencies within blocks. Our experiments remain synthetic, with sequences up to $D=32$ for tokenwise models and $D=16$ for block models. Random source distributions can in principle escape the deterministic-source product floor through mixtures of products, motivating future work on uniform or learned source couplings.

\small
\bibliographystyle{plainnat}
\bibliography{references}

\clearpage

\appendix

\appendix

\section{Additional Theory and Proofs}
\label{app:proofs}

\subsection{Equivalent Integral Form of the Discrete MeanFlow Identity}
\label{app:integral_form}

The Discrete MeanFlow identity also admits an equivalent integral form.
Integrating the Kolmogorov forward equation from $r$ to $t$ and dividing by
$t-r$ gives
\[
\bar{u}(y,x,r,t)
=
\frac{1}{t-r}
\int_r^t
\sum_{z\in\mathcal{S}}
u_\tau(y,z)\,K_{r,\tau}(z,x)\,d\tau.
\]
Thus, $\bar{u}$ is the time-average of the instantaneous probability flow
from $x$ toward $y$ over the interval $[r,t]$.

\subsection{Data-Conditional Kernel-Residual Equivalence}
\label{app:conditional_kr_proof}

We give the full assumptions and loss definitions used in
Proposition~\ref{prop:conditional_kr}.
Assume that the unconditional process $(X_s)_{s\in[0,1]}$ is Markov,
that the time pair $(r,t)$ is sampled independently of the data and
process with $r<t<1$ almost surely, and that the conditional generators
are uniformly bounded and continuous in time on every compact interval
$[0,T]$ with $T<1$.

Define
\[
F_\theta(y,X_r,r,t)
:=
\partial_t K_\theta(y,X_r,r,t),
\]
and let $w(r,t)\geq0$ depend only on the sampled times. We assume
\[
\mathbb{E}\!\left[
w(r,t)\sum_{y\in\mathcal{S}}
F_\theta(y,X_r,r,t)^2
\right]
<\infty
\]
and
\[
\mathbb{E}\!\left[
w(r,t)\sum_{y\in\mathcal{S}}
u_t^{\mathrm{cond}}(y,X_t;X^\star)^2
\right]
<\infty.
\]

The corresponding marginal-target and conditional-target objectives are
\begin{align*}
\mathcal{L}^{w}_{\mathrm{marg}}(\theta)
&=
\mathbb{E}\!\left[
w(r,t)\sum_{y\in\mathcal{S}}
\left(
F_\theta(y,X_r,r,t)
-
u_t^{\mathrm{marg}}(y,X_t)
\right)^2
\right],\\
\mathcal{L}^{w}_{\mathrm{cond}}(\theta)
&=
\mathbb{E}\!\left[
w(r,t)\sum_{y\in\mathcal{S}}
\left(
F_\theta(y,X_r,r,t)
-
u_t^{\mathrm{cond}}(y,X_t;X^\star)
\right)^2
\right].
\end{align*}

For the linear independent-masking process used in our sequence
experiments, each coordinate can equivalently be assigned an independent
reveal time uniformly distributed on $[0,1]$. Once revealed, it displays
its clean value and remains revealed. Conditional on the current state
$X_t$, the past reveal history provides no additional information needed
to determine the future process. Hence the unconditional masking process
is Markov. For a general mixture of conditional Markov processes, this
property need not hold and must be checked separately.

\subsection{Proof of Proposition~\ref{prop:conditional_kr}}

\begin{proof}
The proof has three steps.

\paragraph{Step 1: The conditional target averages to the marginal target.}
Fix $r<t<1$ and $y\in\mathcal{S}$. For convenience, define
\[
U
:=
u_t^{\mathrm{cond}}(y,X_t;X^\star),
\qquad
M
:=
u_t^{\mathrm{marg}}(y,X_t),
\]
and
\[
F
:=
F_\theta(y,X_r,r,t)
=
\partial_t K_\theta(y,X_r,r,t).
\]

By definition of the marginal generator,
\[
M
=
\mathbb{E}[U\mid X_t,t].
\]

We also need the same identity after conditioning on $X_r$. Since
$X^\star=X_1$ and the unconditional process is Markov, once $X_t$ is
known, the earlier state $X_r$ contains no additional information about
the future endpoint $X^\star$. Thus, for $r<t$,
\[
X^\star \perp X_r \mid X_t.
\]
Because the sampled times $(r,t)$ are independent of the process,
\[
\mathbb{E}[U\mid X_r,X_t,r,t]
=
\mathbb{E}[U\mid X_t,t]
=
M.
\tag{A.1}
\label{eq:cond_mean_simple}
\]

Therefore, the sample-wise conditional generator is an unbiased target
for the marginal generator after conditioning on the information relevant
to the regression problem.

\paragraph{Step 2: The two squared losses differ only by a variance term.}
Consider the conditional squared error for this fixed $y$:
\[
(F-U)^2.
\]
Add and subtract $M$:
\[
F-U=(F-M)+(M-U).
\]
Squaring gives
\[
(F-U)^2
=
(F-M)^2
+
2(F-M)(M-U)
+
(M-U)^2.
\]

Now condition on $(X_r,X_t,r,t)$. The quantities $F$ and $M$ are fixed
under this conditioning. By Eq.~\eqref{eq:cond_mean_simple},
\[
\mathbb{E}[M-U\mid X_r,X_t,r,t]=0.
\]
Hence the cross term vanishes:
\[
\mathbb{E}\!\left[
2(F-M)(M-U)
\mid X_r,X_t,r,t
\right]
=0.
\]

The remaining random term is the conditional variance:
\[
\mathbb{E}\!\left[
(M-U)^2
\mid X_r,X_t,r,t
\right]
=
\operatorname{Var}(U\mid X_r,X_t,r,t).
\]
Using the same conditional independence as above,
\[
\operatorname{Var}(U\mid X_r,X_t,r,t)
=
\operatorname{Var}(U\mid X_t,t).
\]

Therefore,
\[
\mathbb{E}\!\left[
(F-U)^2
\mid X_r,X_t,r,t
\right]
=
(F-M)^2
+
\operatorname{Var}(U\mid X_t,t).
\]

Multiplying by the time weight $w(r,t)$, taking expectation over all
random variables, and summing over $y\in\mathcal{S}$ gives
\begin{align}
\mathcal{L}^{w}_{\mathrm{cond}}(\theta)
&=
\mathcal{L}^{w}_{\mathrm{marg}}(\theta)
\nonumber\\
&\quad+
\mathbb{E}\!\left[
w(r,t)\sum_{y\in\mathcal{S}}
\operatorname{Var}\!\left(
u_t^{\mathrm{cond}}(y,X_t;X^\star)
\mid X_t,t
\right)
\right].
\end{align}

The second term contains no dependence on $\theta$. Thus changing the
model parameters changes both objectives in exactly the same way.
Consequently,
\[
\arg\min_\theta
\mathcal{L}^{w}_{\mathrm{cond}}(\theta)
=
\arg\min_\theta
\mathcal{L}^{w}_{\mathrm{marg}}(\theta).
\]
Whenever differentiation and expectation can be interchanged, the two
objectives also have the same expected gradient with respect to
$\theta$.

\paragraph{Step 3: The common population optimum is the true kernel derivative.}
We next show that $u_s^{\mathrm{marg}}$ is the generator of the
unconditional process for every $s<1$.

Fix $s<1$ and states $x,y\in\mathcal{S}$. Conditional on the clean
endpoint $X^\star$, the process has generator
$u_s^{\mathrm{cond}}$. Therefore, for a sufficiently small $h>0$,
\[
\mathbb{P}(X_{s+h}=y\mid X_s=x,X^\star)
=
\delta_{y,x}
+
h\,u_s^{\mathrm{cond}}(y,x;X^\star)
+
o(h).
\]

Taking expectation over $X^\star$ conditional on $X_s=x$ gives
\begin{align}
\mathbb{P}(X_{s+h}=y\mid X_s=x)
&=
\delta_{y,x}
+
h\,
\mathbb{E}\!\left[
u_s^{\mathrm{cond}}(y,x;X^\star)
\mid X_s=x
\right]
+
o(h)
\\
&=
\delta_{y,x}
+
h\,u_s^{\mathrm{marg}}(y,x)
+
o(h).
\end{align}
Thus $u_s^{\mathrm{marg}}$ is the infinitesimal generator of the
unconditional Markov process.

Its transition kernel therefore satisfies the Kolmogorov forward
equation. For $r<t<1$,
\begin{align}
\partial_t K_{r,t}(y,x)
&=
\sum_{z\in\mathcal{S}}
u_t^{\mathrm{marg}}(y,z)
K_{r,t}(z,x)
\\
&=
\mathbb{E}\!\left[
u_t^{\mathrm{marg}}(y,X_t)
\mid X_r=x
\right].
\end{align}

Finally, under squared-error regression, the population-optimal
prediction of a random target is its conditional expectation. Hence,
over unrestricted measurable predictors $F$,
\[
F^\star(y,x,r,t)
=
\mathbb{E}\!\left[
u_t^{\mathrm{marg}}(y,X_t)
\mid X_r=x
\right].
\]
Combining this with the previous equation gives
\[
F^\star(y,x,r,t)
=
\partial_t K_{r,t}(y,x).
\]

Therefore, replacing the unknown marginal-generator target by the
sample-wise data-conditional generator does not change the population
optimum; it only adds conditional variance to the training objective.
\end{proof}

\subsection{Proof of Proposition~\ref{prop:rb}}
\label{app:proof_rb}

\begin{proof}
For part (i), conditional on $(X_r,X^\star,r,t)$, the samples
$X_t^{(1)},\ldots,X_t^{(M)}$ are identically distributed. Therefore,
\[
\mathbb{E}\!\left[
\bar U_M(y)
\,\middle|\,
X_r,X^\star,r,t
\right]
=
\widetilde U(y).
\]
Taking another conditional expectation over $X^\star$ and using
Proposition~\ref{prop:conditional_kr} gives, for $X_r=x$,
\[
\mathbb{E}\!\left[
\widetilde U(y)
\,\middle|\,
X_r=x,r,t
\right]
=
\mathbb{E}\!\left[
\bar U_M(y)
\,\middle|\,
X_r=x,r,t
\right]
=
\partial_t K_{r,t}(y,x).
\]

For parts (ii) and (iii), consider one coordinate under linear masking
and let $m$ denote the mask state. If $X_r$ is already revealed, it
remains equal to $X^\star$ and the conditional generator target is zero.

Suppose instead that $X_r=m$. The probability that the coordinate is
still masked at time $t$ is
\[
q
=
\mathbb{P}(X_t=m\mid X_r=m,r,t)
=
\frac{1-t}{1-r}.
\]
Define
\[
v(y)
=
\mathbf{1}\{y=X^\star\}
-
\mathbf{1}\{y=m\}.
\]
When $X_t=m$, the conditional generator target is
$v(y)/(1-t)$, while after the coordinate has been revealed the target
is zero. Hence
\[
\widetilde U(y)
=
q\,\frac{v(y)}{1-t}
=
\frac{v(y)}{1-r},
\]
which proves Eq.~\eqref{eq:rb_target}.

To compute the second moment, note that
$\sum_y v(y)^2=2$. Conditional on $X_r=m$,
\[
\mathbb{E}\!\left[
\sum_y U(y)^2
\,\middle|\,
X_r=m,r,t
\right]
=
\frac{2}{(1-r)(1-t)},
\]
while
\[
\sum_y
\widetilde U(y)^2
=
\frac{2}{(1-r)^2}.
\]
For the average of $M$ conditionally independent samples,
\[
\begin{aligned}
\mathbb{E}\!\left[
\sum_y \bar U_M(y)^2
\,\middle|\,
X_r=m,r,t
\right]
&=
\frac{2}{(1-r)^2}
\\
&\quad+
\frac{1}{M}
\left(
\frac{2}{(1-r)(1-t)}
-
\frac{2}{(1-r)^2}
\right).
\end{aligned}
\]
Under linear masking,
$\mathbb{P}(X_r=m\mid r)=1-r$. Multiplying by this probability gives
\[
\mathbb{E}\!\left[
\sum_y \bar U_M(y)^2
\,\middle|\,
r,t
\right]
=
\frac{2}{1-r}
+
\frac{2(t-r)}
{M(1-r)(1-t)},
\]
which proves Eq.~\eqref{eq:second_moment}.

The second term is also the expected conditional variance
\[
\mathbb{E}\!\left[
\sum_y
\operatorname{Var}\!\left(
\bar U_M(y)
\,\middle|\,
X_r,X^\star,r,t
\right)
\,\middle|\,
r,t
\right]
=
\frac{2(t-r)}
{M(1-r)(1-t)}.
\]
Thus it is an irreducible contribution to any squared-error regression
using the finite-$M$ target.
\end{proof}

\subsection{Exact Coordinate-Wise Factorization}
\label{app:factorized_exact}

Suppose the CTMC generator decomposes coordinate-wise as
\[
u_t(y,x)
=
\sum_{d=1}^{D}
u_t^{(d)}(y_d,x_d)
\prod_{d'\neq d}
\delta_{y_{d'},x_{d'}}.
\]
Thus, each jump changes at most one coordinate and its rate depends only
on that coordinate. The corresponding transition kernel factorizes as
\[
K_{r,t}(y,x)
=
\prod_{d=1}^{D}
K_{r,t}^{(d)}(y_d,x_d).
\]

Each coordinate therefore has its own mean discrete rate
$\bar u^{(d)}$ and satisfies the Discrete MeanFlow identity
\[
\begin{aligned}
\bar{u}^{(d)}(y_d,x_d,r,t)
&=
\mathbb{E}\!\left[
u_t^{(d)}(y_d,X_t^{(d)})
\,\middle|\,
X_r^{(d)}=x_d
\right]
\\
&\quad-
(t-r)\,
\partial_t\bar{u}^{(d)}(y_d,x_d,r,t).
\end{aligned}
\]

Independent masking and per-position uniform or absorbing corruption
processes factorize coordinate-wise when conditioned on the clean
endpoint. This conditional factorization does not imply that the
data-marginal reverse kernel factorizes when the data distribution
contains dependencies across coordinates.

\subsection{Proof of Proposition~\ref{prop:factorization_lower_bound}}
\label{app:proof_factorization}

\begin{proof}
Let $J=D/B$. For any block-factorized distribution
\[
q(x)
=
\prod_{b=1}^{J}
q_b(x_{\mathcal{B}_b}),
\]
we have
\begin{align*}
D_{\mathrm{KL}}(p\|q)
&=
\sum_x
p(x)
\log
\frac{p(x)}
{\prod_{b=1}^{J}q_b(x_{\mathcal{B}_b})}
\\
&=
-H(X_{1:D})
-
\sum_{b=1}^{J}
\mathbb{E}_{p_{\mathcal{B}_b}}
\left[
\log q_b(X_{\mathcal{B}_b})
\right].
\end{align*}
For each block,
\[
-\mathbb{E}_{p_{\mathcal{B}_b}}
\left[
\log q_b(X_{\mathcal{B}_b})
\right]
=
H(X_{\mathcal{B}_b})
+
D_{\mathrm{KL}}
\left(
p_{\mathcal{B}_b}\|q_b
\right).
\]
Substituting gives
\[
D_{\mathrm{KL}}(p\|q)
=
\mathrm{TC}_B(p)
+
\sum_{b=1}^{J}
D_{\mathrm{KL}}
\left(
p_{\mathcal{B}_b}\|q_b
\right).
\]
Since every KL term is non-negative, the minimum is attained by
$q_b=p_{\mathcal{B}_b}$ for every block, giving
\[
\min_{q=\prod_b q_b}
D_{\mathrm{KL}}(p\|q)
=
\mathrm{TC}_B(p).
\]

Now suppose $p$ is a stationary first-order Markov chain. Let
\[
h
=
H(X_d\mid X_{d-1}),
\]
which is independent of $d$ by stationarity. Then
\[
H(X_{1:D})
=
H(X_1)+(D-1)h.
\]
Each contiguous block of width $B$ has entropy
\[
H(X_{\mathcal{B}_b})
=
H(X_1)+(B-1)h.
\]
Therefore,
\begin{align*}
\mathrm{TC}_B(p)
&=
\frac{D}{B}
\left[
H(X_1)+(B-1)h
\right]
-
\left[
H(X_1)+(D-1)h
\right]
\\
&=
\left(\frac{D}{B}-1\right)
\left[
H(X_1)-h
\right]
\\
&=
\left(\frac{D}{B}-1\right)
I(X_{d-1};X_d).
\end{align*}
This proves the result.
\end{proof}

\subsection{Blockwise Rao-Blackwellized Target}
\label{app:block_rb}

Consider one block and condition on $(X_r,X^\star,r,t)$.
Suppose $m$ coordinates in the block are masked in $X_r$. Each such
coordinate independently reveals its clean value between $r$ and $t$
with probability
\[
p_{r,t}
=
\frac{t-r}{1-r},
\]
and remains masked with probability $1-p_{r,t}$.

For a feasible destination $y_{\mathcal{B}}$, let $S$ denote the subset
of the $m$ masked coordinates that are revealed in
$y_{\mathcal{B}}$, and let $k=|S|$. The conditional block transition
probability is
\[
K^{\mathrm{cond}}_{r,t}
(y_{\mathcal{B}}\mid X_{r,\mathcal{B}},X^\star)
=
p_{r,t}^{\,k}
(1-p_{r,t})^{m-k}.
\]
Since
\[
\partial_t p_{r,t}
=
\frac{1}{1-r},
\]
differentiating gives
\begin{align*}
\partial_t K^{\mathrm{cond}}_{r,t}
&=
\frac{
p_{r,t}^{\,k-1}
(1-p_{r,t})^{m-k-1}
\bigl(k-mp_{r,t}\bigr)
}{1-r}.
\end{align*}
Using
\[
1-p_{r,t}
=
\frac{1-t}{1-r},
\]
we obtain
\[
\partial_t K^{\mathrm{cond}}_{r,t}
=
\frac{
p_{r,t}^{\,k-1}
(1-p_{r,t})^{m-k}
\bigl(k-mp_{r,t}\bigr)
}{1-t}.
\]
By the conditional Kolmogorov forward equation, this derivative is the
conditional expectation of the block generator target given
$(X_r,X^\star,r,t)$, and hence is the Rao-Blackwellized block target in
Eq.~\eqref{eq:block_rb_target}.

If no coordinate in the block is masked at time $r$, then $m=0$ and the
block remains fixed, so the target is zero. Destinations inconsistent
with the already revealed coordinates or with $X^\star$ also have zero
conditional probability and zero target.

\section{Additional Results}
\label{app:stage1}

\subsection{Full Kernel Heatmaps}

\begin{figure}[ht]
    \centering
    \includegraphics[width=\textwidth]{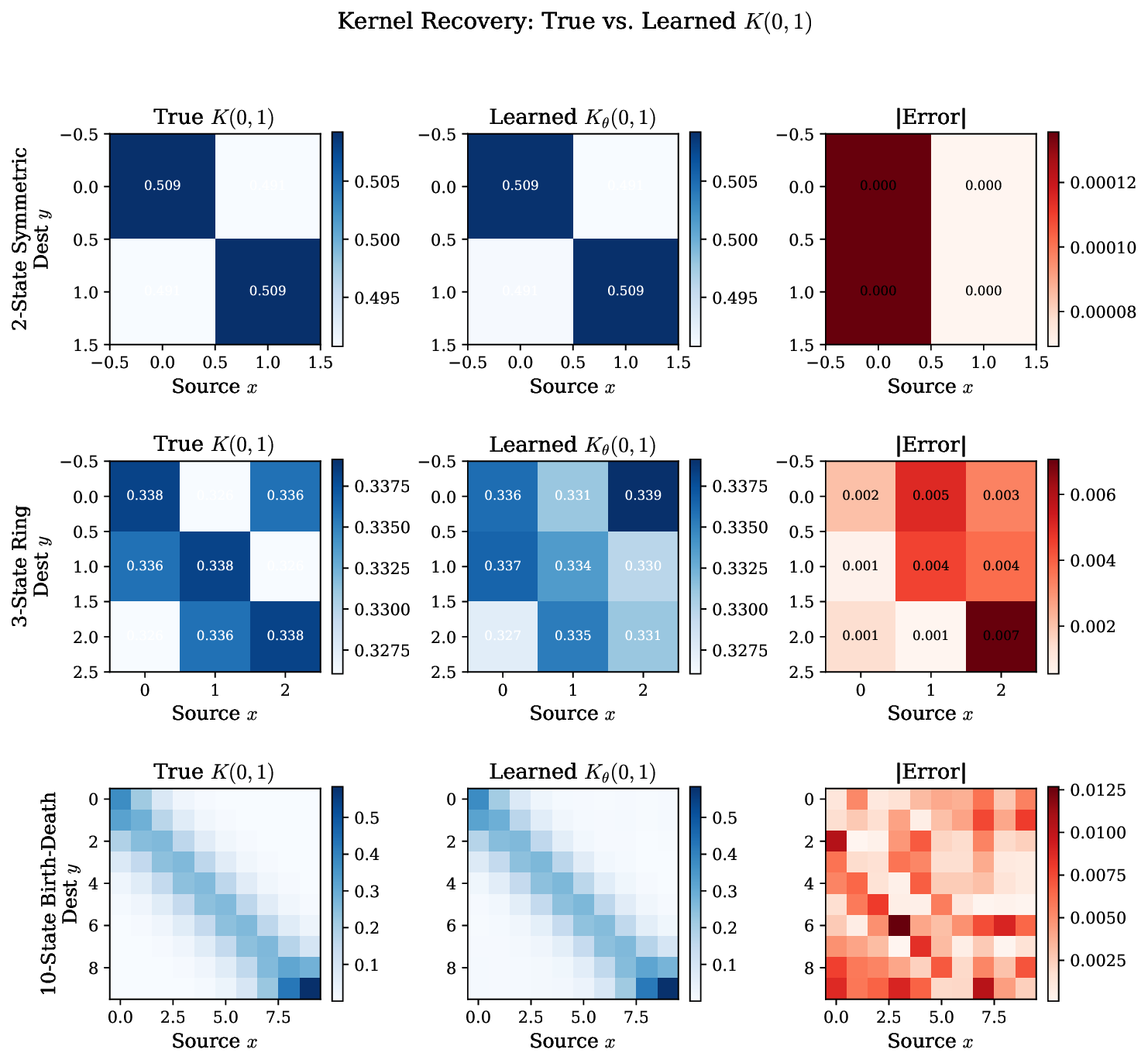}
    \caption{%
        \textbf{Kernel heatmaps for all three CTMCs at $(r,t) = (0,1)$.}
        Each row shows the true kernel (left), learned kernel (middle),
        and absolute error (right).
        Top: 2-state symmetric.
        Middle: 3-state ring.
        Bottom: 10-state birth--death.
    }
    \label{fig:kernel_heatmaps_full}
\end{figure}

Figure~\ref{fig:kernel_heatmaps_full} shows the true kernel, learned
kernel, and absolute error for all three CTMCs evaluated at $(r, t) =
(0, 1)$.
In each case, the learned kernel is visually indistinguishable from the
ground truth, and the error matrices are near zero.

\subsection{Kernel Error Across Time Intervals}
\label{app:error_landscape}

\begin{figure}[t]
    \centering
    \includegraphics[width=\linewidth]{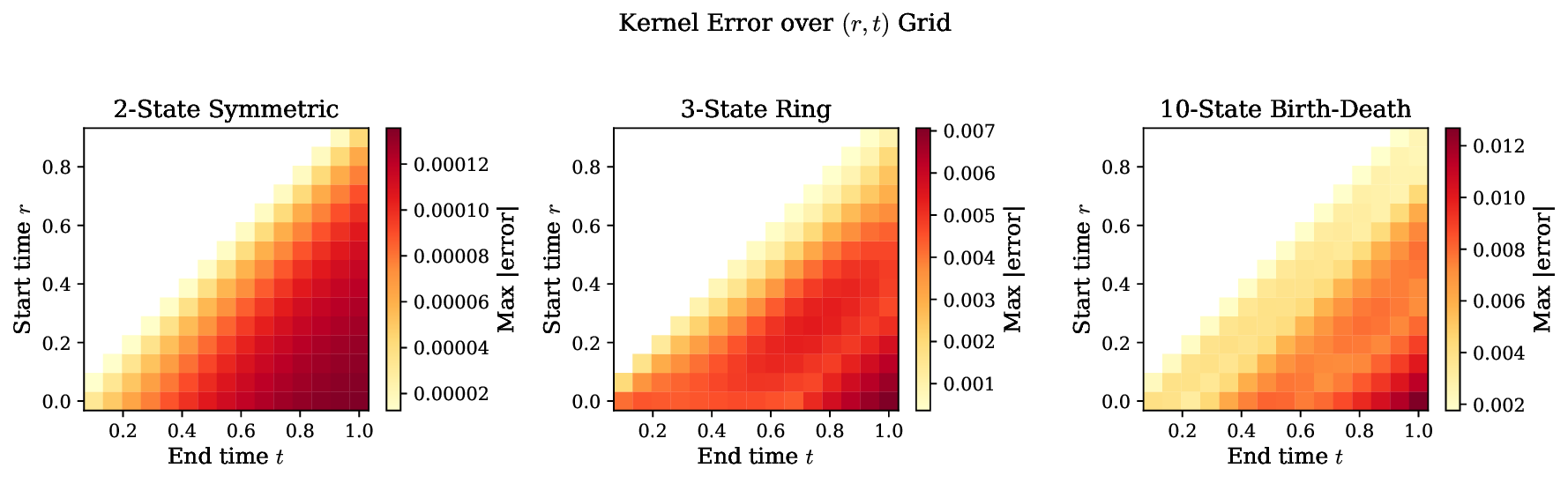}
    \caption{%
        \textbf{Kernel error over the $(r,t)$ grid.}
        Each cell shows the maximum entrywise error
        $\max_{x,y}|K_\theta - K|$ at a given $(r,t)$ pair.
        Error is zero along the diagonal $r=t$ and grows smoothly with
        interval length $t-r$.
        Maximum errors: $1.4\times10^{-4}$, $7.1\times10^{-3}$,
        and $1.3\times10^{-2}$.
    }
    \label{fig:error_landscape}
\end{figure}

The learned kernels remain accurate across the full $(r,t)$ grid, with
zero boundary error at $r=t$ from the boundary-by-construction
parameterization.

\subsection{Training Convergence}

\begin{figure}[t]
    \centering
    \includegraphics[width=\textwidth]{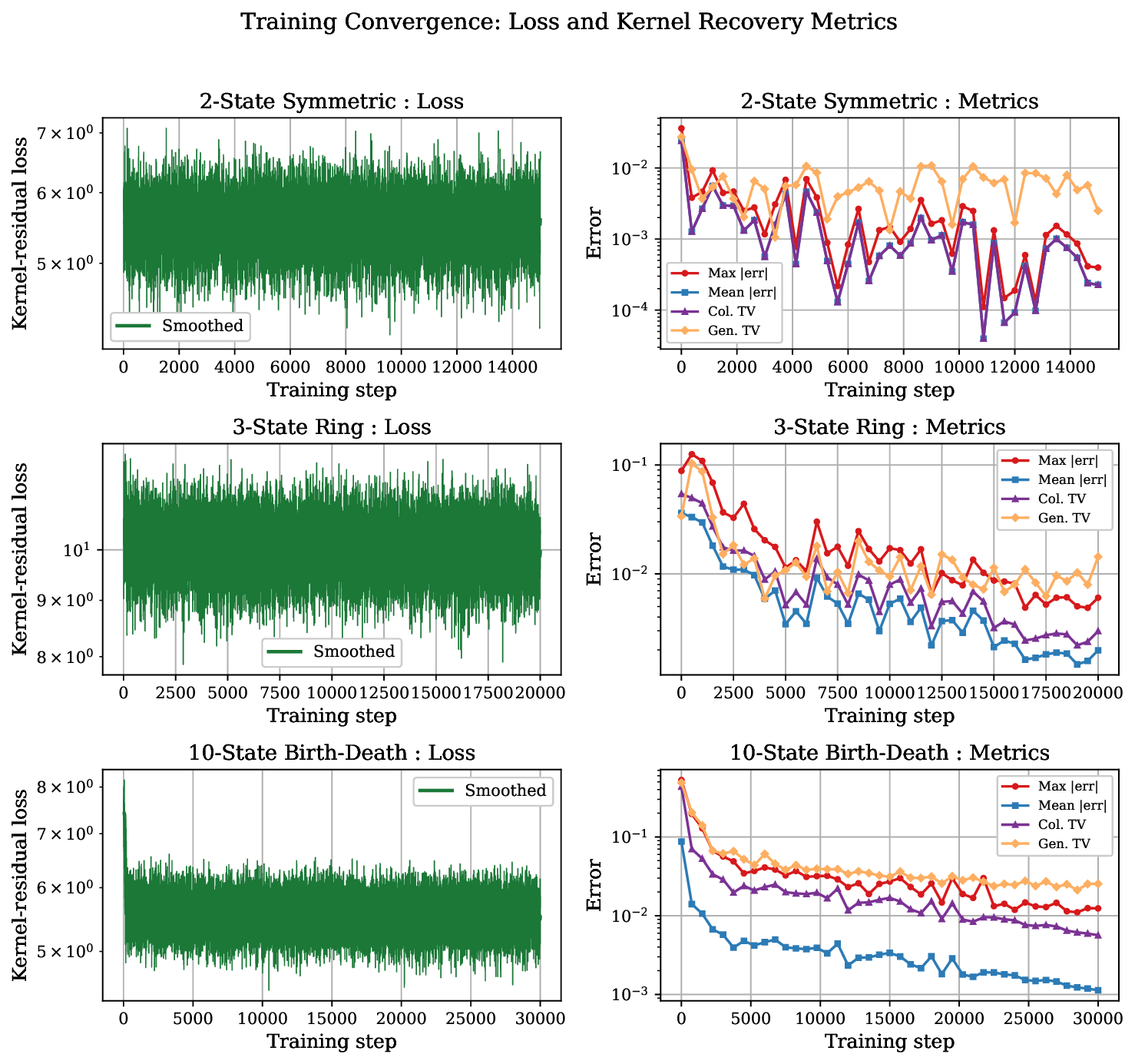}
    \caption{%
        \textbf{Training convergence for Stage~I.}
        Left column: kernel-residual loss over training steps.
        Right column: evaluation metrics (max kernel error, mean kernel
        error, column TV, generation TV) computed periodically during
        training.
    }
    \label{fig:training_convergence}
\end{figure}

Figure~\ref{fig:training_convergence} shows the training loss and
evaluation metrics over the course of training for all three CTMCs.
The loss converges to a stable plateau determined by the irreducible
variance of the stochastic training target.
All error metrics decrease monotonically and stabilize.

\subsection{One-Step Generation Distributions}

\begin{figure}[ht]
    \centering
    \includegraphics[width=\textwidth]{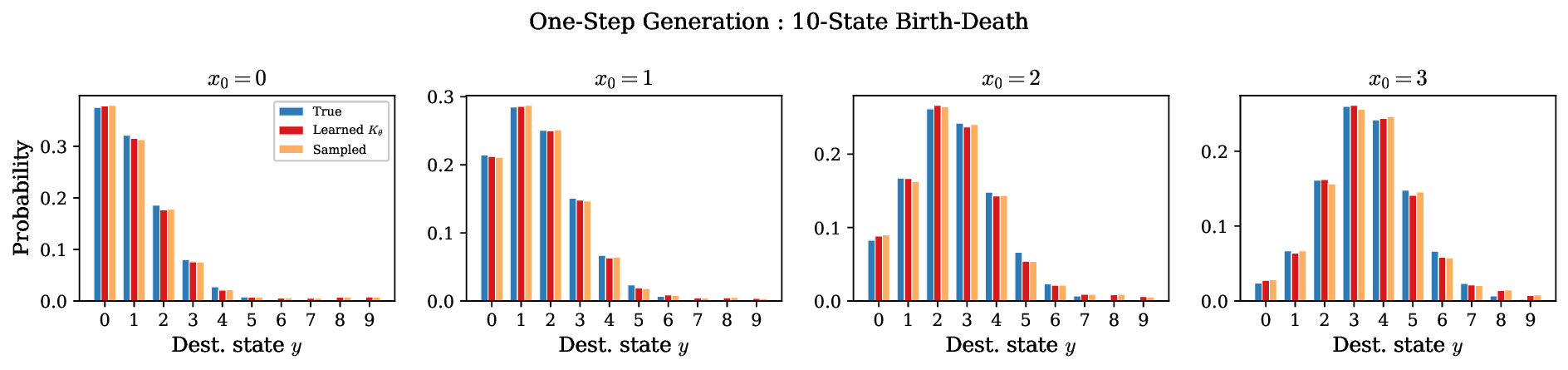}
    \caption{%
        \textbf{One-step generation on the 10-state birth-death chain.}
        For each starting state $x_0$, we compare the true conditional
        distribution (blue), the learned kernel $K_\theta(\cdot, x_0, 0, 1)$
        (red), and the empirical distribution from 10{,}000 one-step samples
        (orange).
    }
    \label{fig:one_step_gen_bd}
\end{figure}

Figure~\ref{fig:one_step_gen_bd} compares the true conditional
distribution, the learned kernel output, and the empirical distribution
of one-step generated samples for the 10-state birth-death chain.
For each starting state $x_0 \in \{0, 1, 2, 3\}$, the three
distributions are closely matched, confirming that a single categorical
draw from $K_\theta(\cdot, x_0, 0, 1)$ faithfully reproduces the true
conditional distribution.

\subsection{Corrected Sequence Objective Results}
\label{app:corrected_sequence_results}

Table~\ref{tab:tv_exact_full} reports one-step marginal TV using the
exact predicted marginals. Increasing the number of endpoint samples
reduces the error of the kernel-residual estimator, Rao-Blackwellization
removes the finite-sample endpoint singularity, and removing the
schedule-induced residual weight gives a further improvement.



\begin{table}[ht]
\centering
\caption{%
\textbf{One-step marginal TV under the corrected sequence setup.}
Exact marginal TV is reported; lower is better. All values are averaged over three seeds.
}
\label{tab:tv_exact_full}

\footnotesize
\setlength{\tabcolsep}{4pt}
\renewcommand{\arraystretch}{0.95}


\begin{tabular}{@{}lcccccc@{}}
\toprule
\textbf{Data}
& \multicolumn{3}{c}{\textbf{Bigram}}
& \multicolumn{3}{c}{\textbf{Independent}} \\
\cmidrule(lr){2-4}
\cmidrule(lr){5-7}

$(|\mathcal{V}|, D)$
& $(4, 8)$
& $(8, 16)$
& $(16, 32)$
& $(4, 8)$
& $(8, 16)$
& $(16, 32)$ \\
\midrule

KR ($M{=}1$)
& 0.017 & 0.023 & 0.033
& 0.065 & 0.067 & 0.060 \\

KR ($M{=}4$)
& 0.009 & 0.019 & 0.028
& 0.019 & 0.028 & 0.045 \\

KR ($M{=}16$)
& 0.009 & 0.017 & 0.031
& 0.009 & 0.019 & 0.037 \\

KR-RB ($M{=}\infty$)
& 0.009 & 0.016 & 0.028
& 0.007 & 0.015 & 0.032 \\

\addlinespace[2pt]

KR ($M{=}1$), unw.
& 0.005 & 0.008 & 0.010
& 0.007 & 0.013 & 0.025 \\

KR-RB, unweighted
& 0.002 & 0.007 & 0.005
& 0.003 & 0.007 & 0.014 \\

\addlinespace[2pt]

WCKR ($M{=}1$)
& 0.004 & 0.007 & 0.010
& 0.005 & 0.010 & 0.019 \\

WCKR-RB
& 0.003 & 0.006 & 0.007
& 0.004 & 0.008 & 0.016 \\

\addlinespace[2pt]

KR (oracle marg.)
& 0.007 & 0.008 & 0.009
& 0.058 & 0.055 & 0.055 \\

PR (Brier)
& 0.003 & 0.005 & 0.006
& 0.004 & 0.011 & 0.020 \\

PR
& 0.003 & 0.005 & 0.005
& 0.002 & 0.002 & 0.004 \\

KCE
& 0.003 & 0.004 & 0.004
& 0.001 & 0.002 & 0.003 \\

\bottomrule
\end{tabular}

\end{table}

For completeness, Table~\ref{tab:tv_sampled_full} reports the corresponding
results when the marginals are estimated from generated samples rather than
computed exactly.



\begin{table}[ht]
\centering
\caption{%
\textbf{Sampled one-step marginal TV under the corrected sequence setup.}
}
\label{tab:tv_sampled_full}

\footnotesize
\setlength{\tabcolsep}{4.2pt}
\renewcommand{\arraystretch}{0.95}


\begin{tabular}{@{}lcccccc@{}}
\toprule
\textbf{Data}
& \multicolumn{3}{c}{\textbf{Bigram}}
& \multicolumn{3}{c}{\textbf{Independent}} \\
\cmidrule(lr){2-4}
\cmidrule(lr){5-7}

$(|\mathcal{V}|, D)$
& $(4, 8)$
& $(8, 16)$
& $(16, 32)$
& $(4, 8)$
& $(8, 16)$
& $(16, 32)$ \\
\midrule

KR ($M{=}1$)
& 0.018 & 0.027 & 0.039
& 0.065 & 0.067 & 0.060 \\

KR ($M{=}4$)
& 0.012 & 0.023 & 0.035
& 0.019 & 0.029 & 0.047 \\

KR ($M{=}16$)
& 0.012 & 0.021 & 0.038
& 0.011 & 0.020 & 0.040 \\

KR-RB ($M{=}\infty$)
& 0.013 & 0.020 & 0.035
& 0.008 & 0.017 & 0.034 \\

\addlinespace[2pt]

KR ($M{=}1$), unw.
& 0.009 & 0.016 & 0.023
& 0.009 & 0.016 & 0.028 \\

KR-RB, unweighted
& 0.008 & 0.015 & 0.022
& 0.005 & 0.010 & 0.018 \\

\addlinespace[2pt]

WCKR ($M{=}1$)
& 0.008 & 0.015 & 0.023
& 0.006 & 0.013 & 0.023 \\

WCKR-RB
& 0.009 & 0.015 & 0.022
& 0.005 & 0.011 & 0.020 \\

\addlinespace[2pt]

KR (oracle marg.)
& 0.011 & 0.016 & 0.023
& 0.059 & 0.056 & 0.056 \\

PR (Brier)
& 0.008 & 0.015 & 0.022
& 0.007 & 0.014 & 0.023 \\

PR
& 0.008 & 0.014 & 0.021
& 0.004 & 0.007 & 0.011 \\

KCE
& 0.008 & 0.015 & 0.021
& 0.004 & 0.007 & 0.010 \\

\bottomrule
\end{tabular}

\end{table}

Token accuracy is reported in Table~\ref{tab:token_acc_full}.



\begin{table}[ht]
\centering
\caption{%
\textbf{One-step token accuracy under the corrected sequence setup.}
}
\label{tab:token_acc_full}

\footnotesize
\setlength{\tabcolsep}{4.2pt}
\renewcommand{\arraystretch}{0.95}


\begin{tabular}{@{}lcccccc@{}}
\toprule
\textbf{Data}
& \multicolumn{3}{c}{\textbf{Bigram}}
& \multicolumn{3}{c}{\textbf{Independent}} \\
\cmidrule(lr){2-4}
\cmidrule(lr){5-7}

$(|\mathcal{V}|, D)$
& $(4, 8)$
& $(8, 16)$
& $(16, 32)$
& $(4, 8)$
& $(8, 16)$
& $(16, 32)$ \\
\midrule

KR ($M{=}1$)
& 0.362 & 0.241 & 0.079
& 0.747 & 0.574 & 0.434 \\

KR ($M{=}4$)
& 0.365 & 0.241 & 0.079
& 0.787 & 0.605 & 0.442 \\

KR ($M{=}16$)
& 0.365 & 0.242 & 0.079
& 0.798 & 0.614 & 0.448 \\

KR-RB ($M{=}\infty$)
& 0.367 & 0.240 & 0.079
& 0.796 & 0.614 & 0.447 \\

\addlinespace[2pt]

KR ($M{=}1$), unw.
& 0.366 & 0.242 & 0.080
& 0.799 & 0.615 & 0.447 \\

KR-RB, unweighted
& 0.366 & 0.242 & 0.078
& 0.798 & 0.616 & 0.449 \\

\addlinespace[2pt]

WCKR ($M{=}1$)
& 0.365 & 0.242 & 0.080
& 0.799 & 0.615 & 0.448 \\

WCKR-RB
& 0.366 & 0.243 & 0.079
& 0.799 & 0.616 & 0.449 \\

\addlinespace[2pt]

KR (oracle marg.)
& 0.364 & 0.240 & 0.079
& 0.753 & 0.585 & 0.435 \\

PR (Brier)
& 0.365 & 0.242 & 0.080
& 0.798 & 0.615 & 0.449 \\

PR
& 0.365 & 0.242 & 0.080
& 0.800 & 0.619 & 0.452 \\

KCE
& 0.365 & 0.241 & 0.080
& 0.800 & 0.619 & 0.452 \\

\bottomrule
\end{tabular}

\end{table}

\subsection{Target Second-Moment Verification}
\label{app:second_moment_full}

\begin{figure}[ht]
    \centering
    \includegraphics[width=\textwidth]{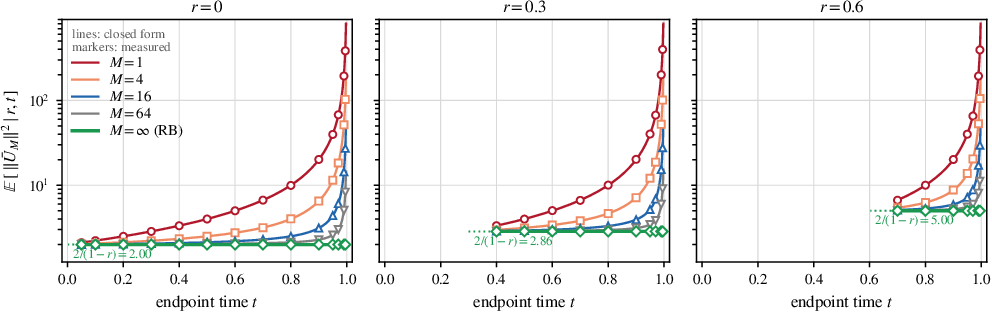}
    \caption{%
    \textbf{Target second moment for three starting times.}
    Closed-form predictions from Eq.~\eqref{eq:second_moment} and measured
    values agree closely over the full grid.
    }
    \label{fig:second_moment_all_r}
\end{figure}

Figure~\ref{fig:second_moment_all_r} verifies
Eq.~\eqref{eq:second_moment} for $r\in\{0,0.3,0.6\}$ and
$M\in\{1,4,16,64,\infty\}$. Lines are the closed-form prediction and
markers are empirical measurements from $20{,}000$ corrupted samples per
cell. The finite-$M$ curves retain the $1/(1-t)$ singularity, while the
Rao-Blackwellized target remains at $2/(1-r)$.

\subsection{Joint Structure and Block Factorization}
\label{app:block_results}

Table~\ref{tab:joint_structure_full} decomposes the one-step forward KL
on the correlated bigram tasks. The total-correlation term accounts for
essentially all joint divergence even when marginal prediction improves.

\begin{table}[ht]
\centering
\caption{%
\textbf{Joint one-step structure under tokenwise factorization.}
The irreducible fraction is the fraction of total KL explained by
$\mathrm{TC}(p)$.
}
\label{tab:joint_structure_full}
\scriptsize
\resizebox{\textwidth}{!}{%
\begin{tabular}{@{}llccccccc@{}}
\toprule
$(|\mathcal{V}|, D)$ & Objective & $\mathrm{TC}(p)$ & $\sum_d \mathrm{KL}(p_d\|q_d)$ & $\mathrm{KL}(p\|q)$ & irred.\ frac. & pair TV & $I_p$ & $I_q$ \\
\midrule
$(4, 8)$ & KR ($M{=}1$) & 1.6849 & 0.0077 & 1.6926 & 0.9954 & 0.2628 & 0.2407 & 0.0001 \\
$(4, 8)$ & KR ($M{=}4$) & 1.6849 & 0.0029 & 1.6878 & 0.9983 & 0.2619 & 0.2407 & 0.0001 \\
$(4, 8)$ & KR ($M{=}16$) & 1.6849 & 0.0025 & 1.6873 & 0.9985 & 0.2624 & 0.2407 & 0.0001 \\
$(4, 8)$ & KR-RB ($M{=}\infty$) & 1.6849 & 0.0028 & 1.6877 & 0.9983 & 0.2625 & 0.2407 & 0.0001 \\
$(4, 8)$ & KR ($M{=}1$), unw. & 1.6849 & 0.0010 & 1.6858 & 0.9994 & 0.2620 & 0.2407 & 0.0001 \\
$(4, 8)$ & KR-RB, unweighted & 1.6849 & 0.0002 & 1.6850 & 0.9999 & 0.2621 & 0.2407 & 0.0001 \\
$(4, 8)$ & WCKR ($M{=}1$) & 1.6849 & 0.0005 & 1.6854 & 0.9997 & 0.2619 & 0.2407 & 0.0001 \\
$(4, 8)$ & WCKR-RB & 1.6849 & 0.0003 & 1.6852 & 0.9998 & 0.2620 & 0.2407 & 0.0001 \\
$(4, 8)$ & KR (oracle marg.) & 1.6849 & 0.0014 & 1.6863 & 0.9991 & 0.2623 & 0.2407 & 0.0001 \\
$(4, 8)$ & PR (Brier) & 1.6849 & 0.0003 & 1.6851 & 0.9998 & 0.2630 & 0.2407 & 0.0001 \\
$(4, 8)$ & PR & 1.6849 & 0.0003 & 1.6851 & 0.9998 & 0.2630 & 0.2407 & 0.0001 \\
$(4, 8)$ & KCE & 1.6849 & 0.0002 & 1.6851 & 0.9999 & 0.2631 & 0.2407 & 0.0001 \\
$(8, 16)$ & KR ($M{=}1$) & 10.8002 & 0.0330 & 10.8332 & 0.9970 & 0.4957 & 0.7200 & 0.0005 \\
$(8, 16)$ & KR ($M{=}4$) & 10.8002 & 0.0227 & 10.8229 & 0.9979 & 0.4958 & 0.7200 & 0.0005 \\
$(8, 16)$ & KR ($M{=}16$) & 10.8002 & 0.0190 & 10.8191 & 0.9982 & 0.4933 & 0.7200 & 0.0005 \\
$(8, 16)$ & KR-RB ($M{=}\infty$) & 10.8002 & 0.0162 & 10.8164 & 0.9985 & 0.4984 & 0.7200 & 0.0005 \\
$(8, 16)$ & KR ($M{=}1$), unw. & 10.8002 & 0.0038 & 10.8040 & 0.9996 & 0.4944 & 0.7200 & 0.0005 \\
$(8, 16)$ & KR-RB, unweighted & 10.8002 & 0.0020 & 10.8022 & 0.9998 & 0.4971 & 0.7200 & 0.0005 \\
$(8, 16)$ & WCKR ($M{=}1$) & 10.8002 & 0.0029 & 10.8031 & 0.9997 & 0.4947 & 0.7200 & 0.0005 \\
$(8, 16)$ & WCKR-RB & 10.8002 & 0.0022 & 10.8024 & 0.9998 & 0.4961 & 0.7200 & 0.0005 \\
$(8, 16)$ & KR (oracle marg.) & 10.8002 & 0.0043 & 10.8045 & 0.9996 & 0.4970 & 0.7200 & 0.0005 \\
$(8, 16)$ & PR (Brier) & 10.8002 & 0.0015 & 10.8017 & 0.9999 & 0.4936 & 0.7200 & 0.0005 \\
$(8, 16)$ & PR & 10.8002 & 0.0014 & 10.8016 & 0.9999 & 0.4937 & 0.7200 & 0.0005 \\
$(8, 16)$ & KCE & 10.8002 & 0.0010 & 10.8012 & 0.9999 & 0.4939 & 0.7200 & 0.0005 \\
$(16, 32)$ & KR ($M{=}1$) & 27.2353 & 0.1298 & 27.3651 & 0.9953 & 0.5404 & 0.8786 & 0.0023 \\
$(16, 32)$ & KR ($M{=}4$) & 27.2353 & 0.0895 & 27.3248 & 0.9967 & 0.5392 & 0.8786 & 0.0023 \\
$(16, 32)$ & KR ($M{=}16$) & 27.2353 & 0.1031 & 27.3384 & 0.9962 & 0.5384 & 0.8786 & 0.0023 \\
$(16, 32)$ & KR-RB ($M{=}\infty$) & 27.2353 & 0.0887 & 27.3240 & 0.9968 & 0.5371 & 0.8786 & 0.0022 \\
$(16, 32)$ & KR ($M{=}1$), unw. & 27.2353 & 0.0113 & 27.2466 & 0.9996 & 0.5373 & 0.8786 & 0.0023 \\
$(16, 32)$ & KR-RB, unweighted & 27.2353 & 0.0031 & 27.2383 & 0.9999 & 0.5374 & 0.8786 & 0.0023 \\
$(16, 32)$ & WCKR ($M{=}1$) & 27.2353 & 0.0112 & 27.2465 & 0.9996 & 0.5371 & 0.8786 & 0.0023 \\
$(16, 32)$ & WCKR-RB & 27.2353 & 0.0064 & 27.2417 & 0.9998 & 0.5374 & 0.8786 & 0.0023 \\
$(16, 32)$ & KR (oracle marg.) & 27.2353 & 0.0114 & 27.2467 & 0.9996 & 0.5379 & 0.8786 & 0.0023 \\
$(16, 32)$ & PR (Brier) & 27.2353 & 0.0044 & 27.2396 & 0.9998 & 0.5372 & 0.8786 & 0.0022 \\
$(16, 32)$ & PR & 27.2353 & 0.0025 & 27.2378 & 0.9999 & 0.5371 & 0.8786 & 0.0022 \\
$(16, 32)$ & KCE & 27.2353 & 0.0016 & 27.2368 & 0.9999 & 0.5371 & 0.8786 & 0.0022 \\
\bottomrule
\end{tabular}

}
\end{table}

\begin{figure}[ht]
    \centering
    \includegraphics[width=\textwidth]{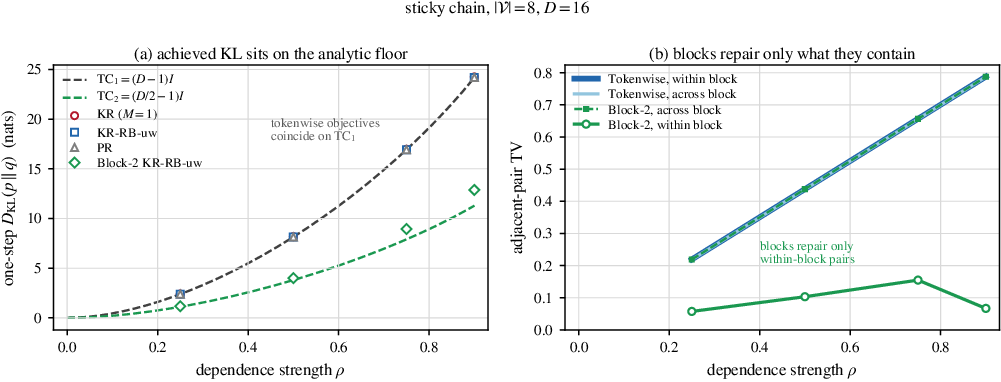}
    \caption{%
    \textbf{Block factorization on the larger sticky-chain setting.}
    Results for $|\mathcal V|=8$ and $D=16$ reproduce the behavior in
    Figure~\ref{fig:block_floor}.
    }
    \label{fig:block_floor_v8d16}
\end{figure}

Figure~\ref{fig:block_floor_v8d16} repeats the block-factorization
experiment at $|\mathcal V|=8$ and $D=16$. The tokenwise objectives again
lie on the analytic product floor, while Block-2 lowers the floor and
recovers dependence within blocks without changing dependence across
block boundaries.

\subsection{Earlier Diagnostic Sequence Experiments}
\label{app:earlier_diagnostics}

The following experiments were conducted during the initial analysis of
Discrete MeanFlow using the true per-position data marginals in place of
the sample-wise conditional generator and an exponential boundary
schedule. They motivated the variance and factorization analyses in the
revised paper, but their numerical values are not directly comparable to
the corrected sequence experiments in
Appendix~\ref{app:corrected_sequence_results}. We retain them as
diagnostic evidence.

\paragraph{Objective comparison.}
Under this earlier setup, posterior regression and kernel cross-entropy
outperformed the kernel-residual objective on per-position marginal TV.
These results motivated the subsequent analysis of target variance and
the corrected experiments in
Appendix~\ref{app:corrected_sequence_results}.

\begin{table}[t]
\centering

\caption{%
    \textbf{Training objective comparison under the earlier diagnostic setup.}
    Average TV distance ($\downarrow$) across three seeds for each
    configuration. PR and KCE achieve lower marginal TV than KR in this setup.
}
\label{tab:objective_comparison}

\scriptsize
\resizebox{0.52\linewidth}{!}{%
\begin{tabular}{@{}llc ccc@{}}
\toprule
Data & $(|\mathcal{V}|, D)$ & KR & PR & KCE & KR/PR \\
\midrule
Independent & $(4, 8)$   & 0.027 & 0.006 & 0.006 & $4.7\times$ \\
Independent & $(8, 16)$  & 0.030 & 0.008 & 0.008 & $3.7\times$ \\
Independent & $(16, 32)$ & 0.053 & 0.013 & 0.013 & $4.1\times$ \\
\midrule
Bigram      & $(4, 8)$   & 0.022 & 0.009 & 0.009 & $2.3\times$ \\
Bigram      & $(8, 16)$  & 0.015 & 0.011 & 0.011 & $1.3\times$ \\
Bigram      & $(16, 32)$ & 0.022 & 0.019 & 0.017 & $1.2\times$ \\
\bottomrule
\end{tabular}%
}

\end{table}

\begin{figure}[t]
    \includegraphics[width=\linewidth]{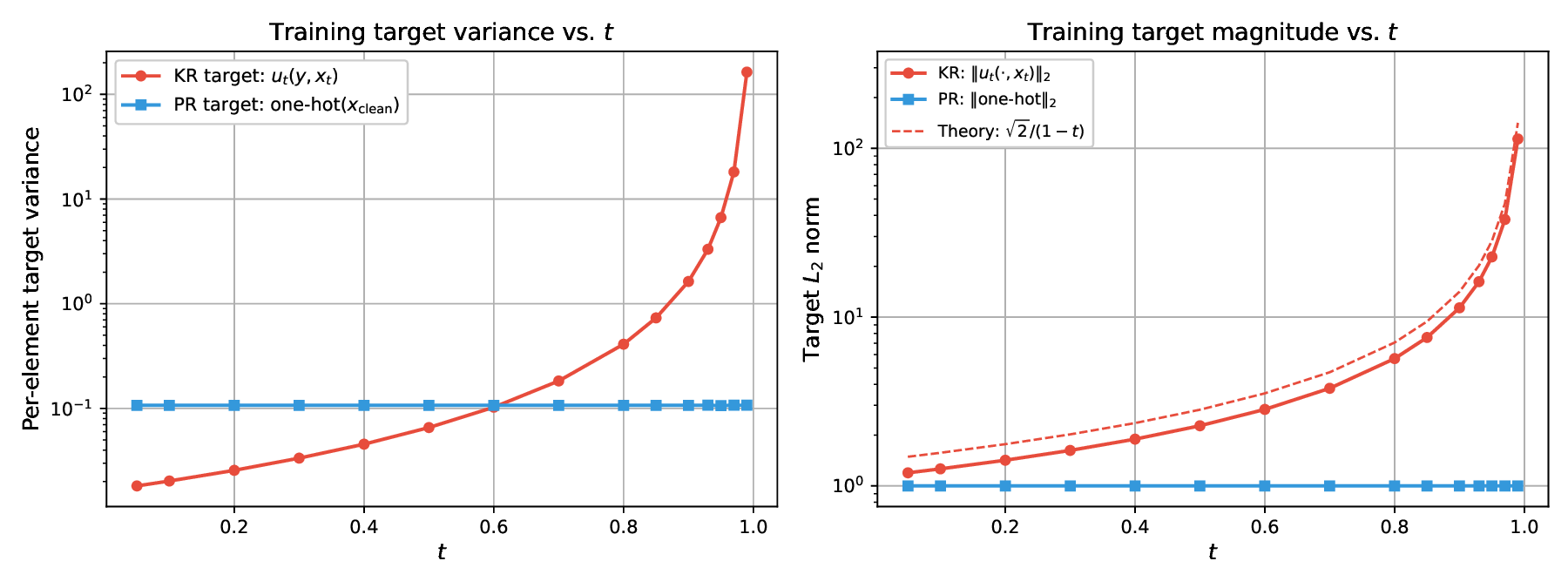}
    \caption{
    Training target variance vs.\ $t$.
    Left: the kernel-residual target has rapidly increasing variance
    near $t=1$, while posterior regression remains bounded.
    Right: conditional on masked positions, the target norm grows
    proportionally to $(1-t)^{-1}$.
    }
    \label{fig:target_variance}
    \vspace{-12pt}
\end{figure}

\paragraph{Initial variance diagnosis.}
The kernel-residual (KR) loss trains $\partial_t K_\theta$ to match the instantaneous generator $u_t(y,x_t)$ at a single sampled endpoint $x_t$, which introduces two sources of high variance. First, masking corruption makes generator entries scale as $1/(1-t)$, diverging as $t \to 1$. Second, a single $x_t$ gives a noisy estimate of the conditional expectation $\mathbb{E}[u_t(y,X_t)\mid X_r=x]$ in the Discrete MeanFlow identity. In contrast, posterior regression uses the clean token as a bounded, deterministic target that directly supervises each position. Thus, estimator variance is one plausible contributor to the KR--PR gap: exact kernel recovery confirms that the identity and parameterization can recover the correct kernel in controlled settings, while the naive single-sample estimator yields high-variance gradients. Figure~\ref{fig:target_variance} directly quantifies this effect. For each $t$, we draw 5000 samples and measure the per-element target variance. The KR target variance grows as $1/(1-t)^2$, reaching over $1500\times$ the posterior-regression variance at $t=0.99$, while the posterior-regression target variance remains constant. The right panel further shows that, conditional on a masked position, the $L_2$ norm of the KR target closely follows the theoretical scaling $\sqrt{2}/(1-t)$. These observations motivated the Rao-Blackwellized analysis and corrected
experiments in the revised paper; see
Proposition~\ref{prop:rb} and
Appendix~\ref{app:corrected_sequence_results}.

\paragraph{Initial factorization diagnosis.}
The earlier bigram experiments also suggested that marginal metrics hide
joint dependence error. With a deterministic fully masked source, the
tokenwise output is a product distribution, so
\[
D_{\mathrm{KL}}(p\|q)
=
\mathrm{TC}(p)
+
\sum_d D_{\mathrm{KL}}(p_d\|q_d).
\]
For first-order Markov data,
$\mathrm{TC}(p)=\sum_{d=2}^{D}I(X_{d-1};X_d)$.
This observation motivated the generalized block-factorization analysis
in Proposition~\ref{prop:factorization_lower_bound}.

\paragraph{Kernel-Residual vs.\ Posterior-Regression}

\begin{figure}[ht]
    \centering
    \includegraphics[width=\textwidth]{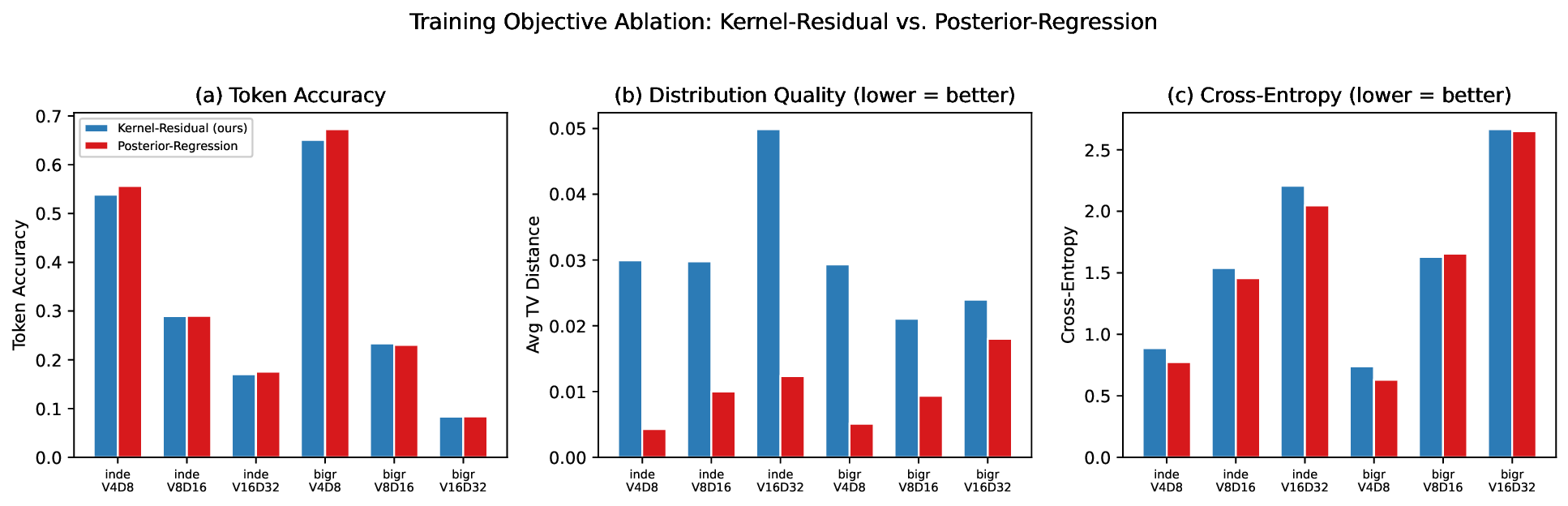}
    \caption{%
        \textbf{Training objective ablation: kernel-residual vs.\
        posterior-regression.}
        (a)~Token accuracy is similar.
        (b)~Posterior regression achieves lower TV distance in all
        configurations.
        (c)~Cross-entropy is also lower for posterior regression.
    }
    \label{fig:s2_ablation_comparison}
\end{figure}

Figure~\ref{fig:s2_ablation_comparison} provides a detailed visual
comparison of the kernel-residual and posterior-regression training
objectives across all six configurations.
Token accuracy is comparable between the two methods, but posterior
regression achieves consistently lower TV distance and cross-entropy,
confirming that direct supervision produces better-calibrated
distributions.

\paragraph{Multi-Step Generation}

\begin{figure}[ht]
    \centering
    \includegraphics[width=\textwidth]{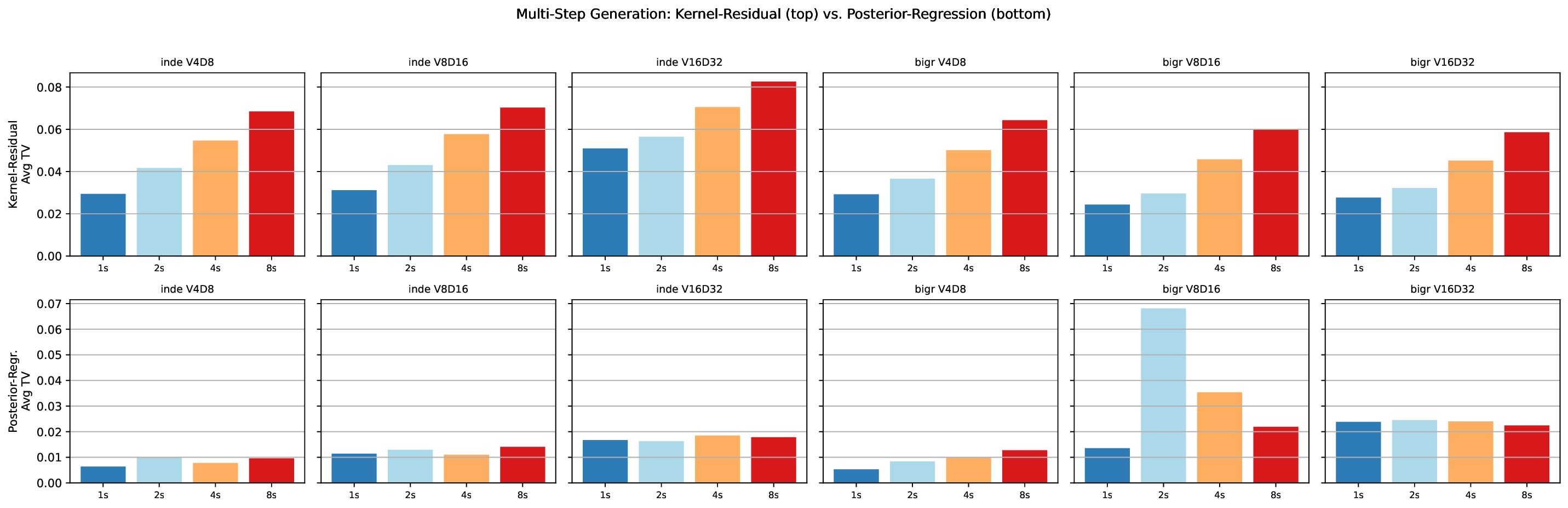}
    \caption{%
        \textbf{Multi-step generation comparison.}
        Top row: kernel-residual.
        Bottom row: posterior-regression.
        Each panel shows average TV distance at 1, 2, 4, and 8
        generation steps.
    }
    \label{fig:s2_ablation_steps}
\end{figure}

Figure~\ref{fig:s2_ablation_steps} compares one-step and multi-step
generation for both training objectives.
The kernel-residual model degrades with additional steps (TV increases
from 1-step to 8-step), while the posterior-regression model remains
stable or slightly improves, indicating that the kernel-residual model
has not learned transition dynamics accurate enough for iterative
composition.

\paragraph{Hybrid Loss Sweep}

\begin{figure}[ht]
    \centering
    \includegraphics[width=\textwidth]{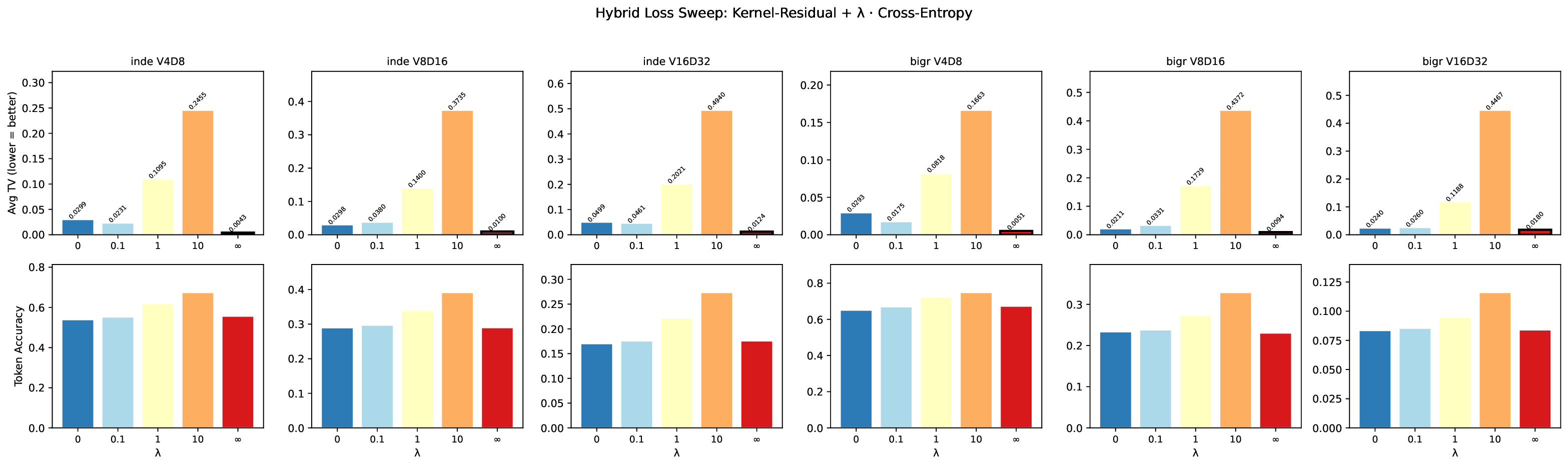}
    \caption{%
        \textbf{Hybrid loss sweep.}
        Top row: average TV distance vs.\ $\lambda$ for each
        configuration.
        Bottom row: token accuracy vs.\ $\lambda$.
        The best TV is consistently achieved at $\lambda = \infty$
        (pure cross-entropy).
    }
    \label{fig:s2_hybrid_sweep}
\end{figure}

We investigated whether combining the kernel-residual and cross-entropy
losses could outperform either alone, using the hybrid objective
$\mathcal{L} = \mathcal{L}_{\mathrm{KR}} + \lambda \cdot
\mathcal{L}_{\mathrm{CE}}$ with $\lambda \in \{0, 0.1, 1, 10, \infty\}$.
Figure~\ref{fig:s2_hybrid_sweep} shows the results.
Pure posterior regression ($\lambda = \infty$) achieves the lowest TV
distance in all configurations.
Intermediate values of $\lambda$ are actively harmful: they increase
token accuracy (the model becomes overconfident) while degrading TV
distance (the predicted distributions are miscalibrated).
Under this earlier diagnostic setup, intermediate hybrid weights do not
improve over either objective alone.

\begin{figure}[ht]
    \centering
    \includegraphics[width=\textwidth]{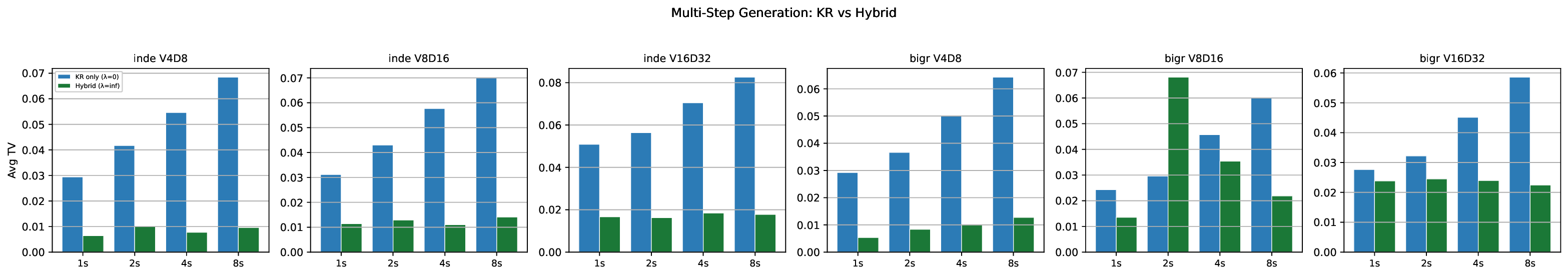}
    \caption{%
        \textbf{Multi-step generation from the hybrid sweep.}
        Kernel-residual (blue) degrades with more steps; posterior
        regression (green) remains stable.
    }
    \label{fig:s2_hybrid_steps}
\end{figure}

Figure~\ref{fig:s2_hybrid_steps} shows the multi-step generation
comparison between pure kernel-residual ($\lambda = 0$) and pure
posterior regression ($\lambda = \infty$) from the hybrid sweep.
The posterior-regression model maintains stable quality across step
counts, while the kernel-residual model degrades significantly.

\paragraph{Step-Count Comparison}

\begin{figure}[ht]
    \centering
    \includegraphics[width=\textwidth]{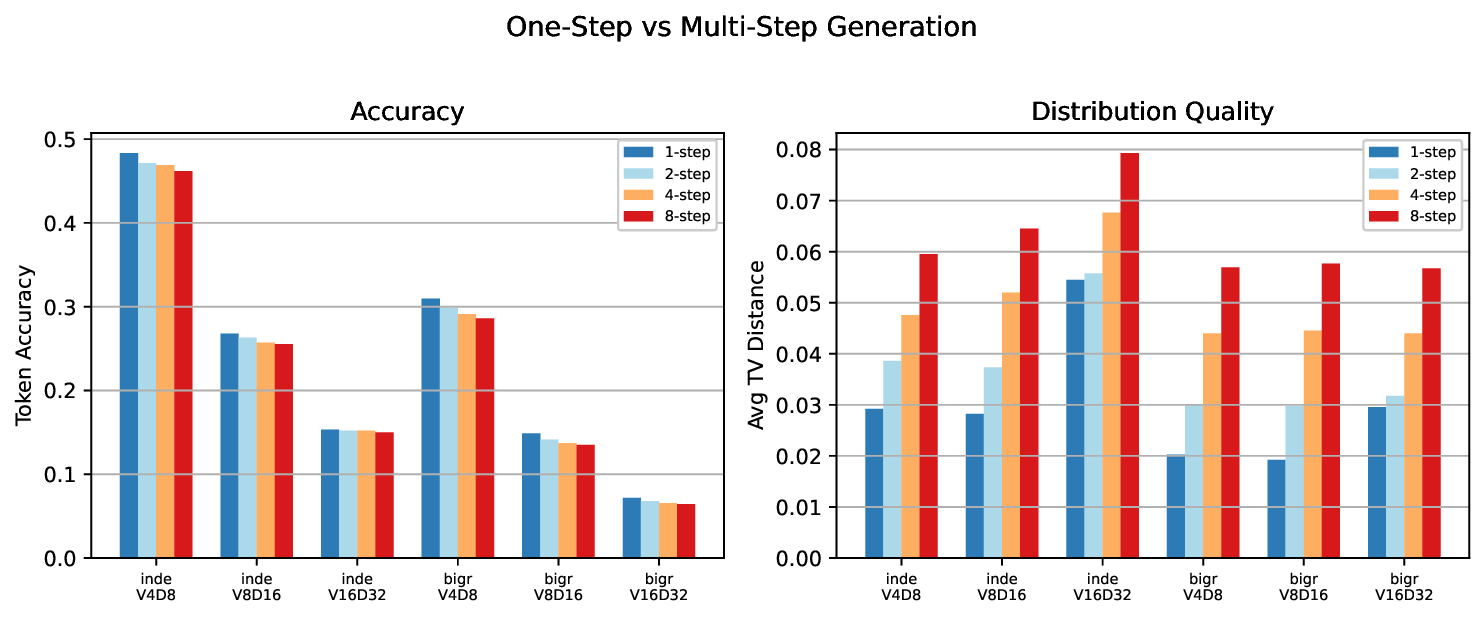}
    \caption{%
        \textbf{Generation quality vs.\ number of sampling steps.}
    }
    \label{fig:s2_step_comparison}
\end{figure}

Figure~\ref{fig:s2_step_comparison} provides an additional view of
generation quality as a function of the number of sampling steps across
all configurations.

\subsection{Design Ablations}

We ablate several design choices on the 3-state ring CTMC, reporting mean kernel TV distance over three seeds.

\paragraph{Boundary treatment ablation.}

\begin{figure}[t]
    \centering
    \includegraphics[width=\linewidth]{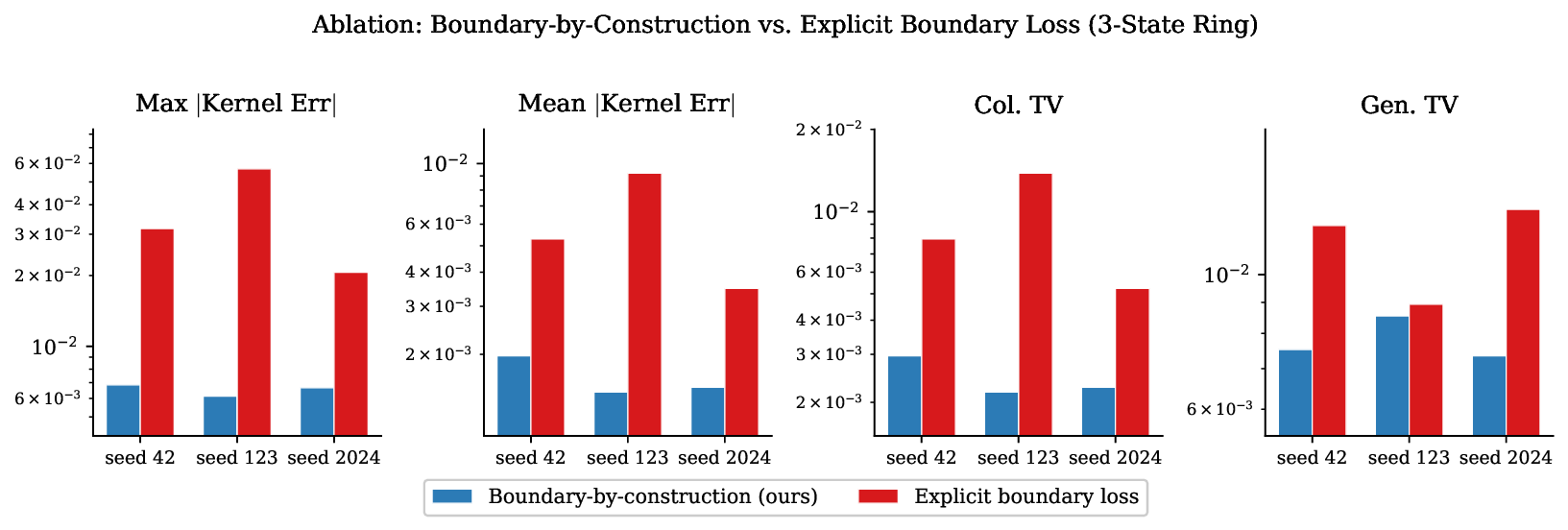}
    \caption{%
        \textbf{Boundary treatment ablation on the 3-state ring.}
        Boundary-by-construction (blue) achieves $3$-$10\times$ lower
        error than an explicit boundary loss (red) across all four
        metrics and all three seeds, under this controlled CTMC setup.
    }
    \label{fig:ablation_boundary}
\end{figure}

The boundary-by-construction parameterization is a key design choice.
To justify it, we compare against an alternative that uses an
unconstrained softmax output and enforces the boundary condition through
an explicit penalty term in the loss with weight $\lambda_{\mathrm{bdy}}
= 10$.
Figure~\ref{fig:ablation_boundary} shows the comparison on the 3-state
ring CTMC across three random seeds.
Boundary-by-construction achieves $3$ -- $10\times$ lower error than the
explicit boundary loss on every metric and every seed.
The advantage is consistent across maximum kernel error, mean kernel
error, column-wise total variation, and one-step generation total
variation. This confirms that enforcing the boundary by construction is more effective than using a penalty loss and avoids tuning a boundary-loss weight.

\paragraph{Fraction of training samples with $r \neq t$.}
The training procedure samples time pairs $(r, t)$ where $r < t$.
Setting $r = t$ for all samples recovers standard instantaneous
generator training and removes the finite-interval structure of Discrete
MeanFlow entirely.
Table~\ref{tab:design_ablations} shows that using $r = t$ for all
samples yields a TV distance of 0.079, roughly $20\times$ worse than
any configuration that includes $r \neq t$ pairs.
This confirms that the two-time conditioning is essential for learning
transition kernels over finite intervals.
Among the nonzero ratios, performance is stable, with the best result
at 100\% (TV = 0.002).

\paragraph{Time sampling distribution.}
Replacing the uniform time sampler with a logit-normal distribution
yields a small improvement (TV = 0.003 vs.\ 0.003), suggesting that
concentrating samples near the boundaries of $[0,1]$ helps but is
not critical.

\begin{table}[ht]
\centering
\caption{%
    \textbf{Design ablations on the 3-state ring.}
    Mean kernel TV distance over 3 seeds.
}
\label{tab:design_ablations}
\small
\begin{tabular}{@{}lc@{\hspace{2em}}lc@{}}
\toprule
\multicolumn{2}{c}{(a) Ratio of $r \neq t$}
    & \multicolumn{2}{c}{(b) Time sampler} \\
\cmidrule(r){1-2} \cmidrule(l){3-4}
Ratio & TV & Sampler & TV \\
\midrule
0\%   & 0.079 & Uniform           & 0.003 \\
25\%  & 0.005 & Logit-$\mathcal{N}(0, 1)$     & 0.003 \\
50\%  & 0.003 & Logit-$\mathcal{N}(-0.4, 1)$  & 0.003 \\
100\% & 0.002 &                   &       \\
\bottomrule
\end{tabular}
\end{table}

\begin{figure}[t]
    \centering
    \includegraphics[
        width=\linewidth
    ]{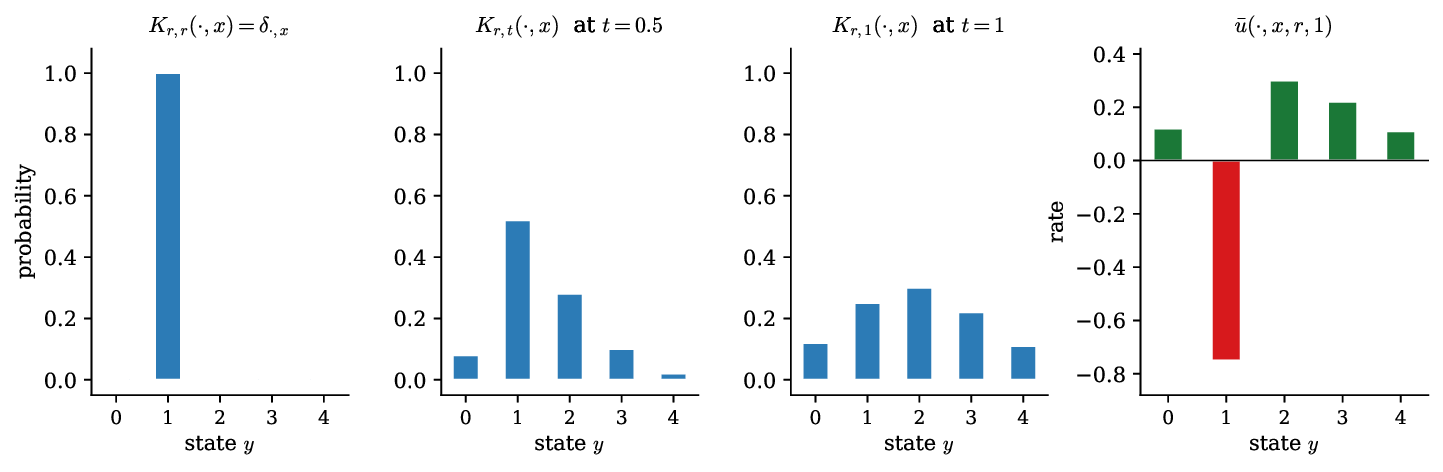}

    \vspace{-3pt}
    \caption{%
        \textbf{The mean discrete rate.}
        From source state $x{=}1$, the transition kernel $K_{r,t}(\cdot,x)$
        spreads probability mass over time.
        The mean discrete rate $\bar{u}$ summarizes this change: positive entries gain
        probability, negative entry at source state loses probability, and
        column sums to zero.
    }
    \label{fig:method}

    \vspace{-6pt}
\end{figure}

\section{Experimental Details}
\label{app:experimental_details}

This appendix provides the full experimental setup needed to reproduce all results in the paper.

\subsection{Boundary Parameterization Details}
\label{app:boundary_param}

The choice of the mixing schedule $\alpha(r,t)$ controls the behavior of
the boundary-by-construction kernel near $t=r$.
From Eq.~\eqref{eq:kernel_param}, the induced mean discrete rate is
\[
\bar{u}_\theta
=
\frac{K_\theta-\delta}{t-r}.
\]
As $t\to r$, its limit is
\[
\lim_{t\to r}\bar{u}_\theta
=
\left.\partial_t\alpha(r,t)\right|_{t=r}
\bigl(q_\theta-\delta\bigr).
\]
Thus, to obtain a nondegenerate finite boundary rate, $\alpha$ should
vanish linearly at $t=r$, with
$\left.\partial_t\alpha(r,t)\right|_{t=r}=c(r)>0$.

For the exact finite-state CTMC experiments, we use
\[
\alpha(r,t)=1-\exp\!\bigl(-c(t-r)\bigr),
\]
where $c=2\max_{x,y,t}|u_t(y,x)|$ in our exact CTMC experiments.

For the linear-masking sequence experiments, we use
\[
\alpha(r,t)
=
\frac{t-r}{1-r+10^{-6}}.
\]
Both schedules vanish linearly as $t\to r$ and therefore satisfy the
required boundary behavior.

\subsection{Training Algorithm and Reference Sampling}
\label{app:training_algorithm}

\begin{algorithm}[ht]
\caption{Discrete MeanFlow training}
\label{alg:train}
\begin{algorithmic}[1]
\Require Reference generator $u_t$; transition kernel $K_{r,t}$
(analytic or simulable); kernel model $K_\theta$
\Repeat
    \State Sample times
    $r\sim\mathrm{Unif}[0,1-\epsilon)$,
    $t\sim\mathrm{Unif}[r+\epsilon,1]$
    \State Sample $x_r\sim p_r$ and
    $x_t\sim K_{r,t}(\cdot,x_r)$
    \State Compute $\partial_tK_\theta(\cdot,x_r,r,t)$
    \State Set $g(y)\gets u_t(y,x_t)$ for all $y\in\mathcal S$
    \State $\mathcal L\gets
    \sum_y\bigl(
    \partial_tK_\theta(y,x_r,r,t)-g(y)
    \bigr)^2$
    \State Update $\theta\gets\theta-\eta\nabla_\theta\mathcal L$
\Until{converged}
\end{algorithmic}
\end{algorithm}

The derivative $\partial_tK_\theta$ is computed by automatic
differentiation with respect to the scalar endpoint time $t$. From
Eq.~\eqref{eq:kernel_param},
\[
\partial_t K_\theta(y,x,r,t)
=
(\partial_t\alpha)
\bigl(q_\theta(y\mid x,r,t)-\delta_{y,x}\bigr)
+
\alpha\,\partial_tq_\theta(y\mid x,r,t).
\]

The expectation in Theorem~\ref{thm:discrete-meanflow} is taken under
the reference CTMC. We therefore draw
$X_t\sim K_{r,t}(\cdot,X_r)$ rather than from the learned kernel
$K_\theta$. Since this sampling distribution is fixed independently of
$\theta$, Eq.~\eqref{eq:loss} gives an unbiased stochastic gradient of
the corresponding population objective. Sampling from $K_\theta$ would
instead define a different objective.

Drawing $X_t\sim K_{r,t}(\cdot,X_r)$ requires the reference transition
kernel to be sampleable. For the masking and other coordinate-wise
corruption processes considered here, conditional sampling factorizes
across positions and costs $O(D)$ per sequence. For a time-homogeneous
finite-state generator $Q$, the transition kernel is
$K_{r,t}=\exp((t-r)Q)$ and can also be precomputed and sampled directly.
More general reference processes without tractable or simulable
transition kernels are outside the scope of this work.

\subsection{Compute Resources}
\label{app:compute}

All experiments were run on a GPU cluster using NVIDIA Tesla P100 GPUs (12\, GB VRAM) and NVIDIA Tesla V100 GPUs (32\,GB VRAM). Every experiment uses a single GPU with no multi-GPU or distributed training. Peak GPU memory usage is approximately 1.5\,GB for exact kernel recovery and approximately 3\,GB for the factorized sequence experiments, so all experiments can run on any modern GPU with at least 4\,GB of memory.

Exact kernel recovery experiments on 2, 3, and 10 state CTMCs each require a single GPU and complete in 5 to 30 minutes depending on the number of states and training iterations.
Factorized sequence task experiments each require a single GPU and complete in 5 to 20 minutes per configuration per training objective.
The boundary ablation, design ablation, and hybrid loss sweep experiments each complete in under one hour on a single GPU.
The target variance measurement in Figure~\ref{fig:target_variance} requires no GPU and runs in under one minute on a CPU.

The total compute for all experiments reported in the paper, including three seeds per configuration, is approximately 50 GPU hours on P100 hardware.
The full research project, including preliminary experiments and hyperparameter exploration not reported in the paper, used approximately 200 GPU hours in total.

\subsection{Exact kernel recovery experiment setup}

We train on three reference CTMCs whose true transition kernels are available in closed form via matrix exponentiation.

\paragraph{Reference processes.}
The 2 state symmetric chain has rate $\lambda = 2$ in both directions.
The 3 state ring has clockwise rate $\lambda_f = 2$ and counterclockwise rate $\lambda_b = 1$.
The 10 state birth death chain has birth rate $b = 1.5$ and death rate $d = 1.0$ with reflecting boundaries.
In all cases the initial distribution $p_0$ is uniform over states.

\paragraph{Model architecture.}
The kernel network is a 4 layer MLP with hidden dimension 128 and GELU activations.
The input is the concatenation of a one hot encoding of the starting state $x$ (dimension $|\mathcal S|$) and sinusoidal embeddings of $r$ and $t$ (dimension 32 each), giving a total input dimension of $|\mathcal S| + 64$.
The output is a vector of $|\mathcal S|$ logits passed through a softmax to produce $q_\theta(y \mid x, r, t)$.
The boundary-by-construction schedule is described in
Appendix~\ref{app:boundary_param}, with
$c=2\max_{x,y,t}|u_t(y,x)|$ for these experiments.

\paragraph{Training.}
We use AdamW with learning rate $3 \times 10^{-4}$, weight decay $10^{-5}$, and cosine annealing to $1\%$ of the initial learning rate.
Batch size is 256.
Times are sampled as $r \sim \mathrm{Uniform}[0, 1 - \epsilon)$ and $t \sim \mathrm{Uniform}[r + \epsilon, 1]$ with $\epsilon = 0.01$.
Training runs for 15{,}000 iterations (2 state), 20{,}000 iterations (3 state), and 30{,}000 iterations (10 state).
Gradients are clipped to a maximum norm of 1.0.

\paragraph{Evaluation.}
Kernel accuracy is evaluated on a $20 \times 20$ grid of $(r, t)$ pairs.
For each grid point, we compute the learned kernel $K_\theta(\cdot, x, r, t)$ for every starting state $x$ and compare against the true kernel $K_{r,t}$ obtained by matrix exponentiation.
One-step generation quality is measured by drawing 5{,}000 samples from $K_\theta(\cdot, x_0, 0, 1)$ for each starting state $x_0$ and comparing the empirical distribution against the true conditional $K_{0,1}(\cdot, x_0)$ using total variation distance.

\subsection{Sequence Task Setup}
\label{app:sequence_setup}

\paragraph{Data sources.}
\emph{Independent categorical} draws each position independently from a
fixed per-position distribution sampled once from a Dirichlet prior with
concentration $0.1$.
\emph{Bigram chain} uses a random row-stochastic matrix $T$ sampled from
a Dirichlet prior with concentration $0.3$, mixed toward its stationary
distribution $\pi$ as
$T_\lambda[a,b]=(1-\lambda)\pi[b]+\lambda T[a,b]$.
The stationary marginals therefore remain fixed as $\lambda$ changes.
\emph{Sticky chain} uses
$P_\rho(j\mid i)=\rho\mathbf 1\{j=i\}+(1-\rho)/|\mathcal V|$ and has
uniform stationary marginals for every $\rho$.

\paragraph{Model and training.}
We use a three-layer Transformer with model dimension $128$, four
attention heads, learned positional embeddings, and sinusoidal embeddings
of $r$ and $t$. The time derivative is assembled as
\[
\partial_tK_\theta
=
(\partial_t\alpha)(q_\theta-\delta)
+
\alpha\,\partial_tq_\theta,
\]
with $\partial_tq_\theta$ computed using a forward-mode
Jacobian--vector product. AdamW uses learning rate $3\times10^{-4}$,
weight decay $10^{-5}$, cosine annealing, and gradient clipping at norm
$1.0$. Batch size is $256$ for kernel objectives and $128$ for PR.
We sample
$r\sim\mathrm{Unif}[0,1-\varepsilon)$ and
$t\sim\mathrm{Unif}[r+\varepsilon,1]$ with $\varepsilon=0.02$.
All reported main results use three seeds.

\paragraph{Evaluation.}
One-step generation begins from the deterministic fully masked input.
For tokenwise models, marginal TV,
$\sum_dD_{\mathrm{KL}}(p_d\|q_d)$, and total KL are therefore computed
directly from the predicted product distribution without sampling noise.
Adjacent-pair TV and generated mutual information are estimated from
$50{,}000$ generated sequences.

\subsection{Boundary Ablation Setup}

The boundary ablation in Figure~\ref{fig:ablation_boundary} compares boundary-by-construction against an unconstrained softmax baseline on the 3 state ring CTMC.
The unconstrained baseline uses the same MLP architecture but without the $\alpha$-mixing and instead adds a boundary penalty $\lambda_{\mathrm{bdy}} \sum_{x} \|K_\theta(\cdot, x, r, r) - \delta_{\cdot,x}\|^2$ to the training loss with $\lambda_{\mathrm{bdy}} = 10$.
All other hyperparameters are identical to the exact kernel recovery experiment setup.
Results are reported over three seeds.

\subsection{Design Ablation Setup}

The design ablations in Table~\ref{tab:design_ablations} are run on the 3 state ring CTMC with the same model and training setup as kernel recovery experiments (15{,}000 iterations, batch size 256).

For the ratio ablation, we vary the fraction of training samples where $r \neq t$ from 0\% to 100\%.
At 0\%, all samples use $r = t$ (with a small offset $\epsilon = 0.01$ for numerical stability), which recovers standard instantaneous generator training.
At 100\%, all samples use independently sampled $r < t$.

For the time sampler ablation, we compare uniform sampling of $t$ against two logit normal distributions.
For logit normal sampling, we draw $z \sim \mathcal{N}(\mu, 1)$ and set $t = \sigma(z)$ where $\sigma$ is the sigmoid function.
The two variants use $\mu = 0$ and $\mu = -0.4$.

\subsection{Target Second-Moment Measurement}

The measurement in Section~\ref{sec:experiments} uses no trained model.
For each $(r,t,M)$ cell with $r\in\{0,0.3,0.6\}$, $t$ on a grid from
$0.05$ to $0.995$, and $M\in\{1,4,16,64,\infty\}$, we draw
$20{,}000$ corrupted samples and average
$\sum_y\bar U_M(y)^2$ over coordinates. We compare the result with
Eq.~\eqref{eq:second_moment}.

\subsection{Earlier Hybrid Loss Sweep Setup}

The hybrid loss experiments in Figures~\ref{fig:s2_hybrid_sweep} and~\ref{fig:s2_hybrid_steps} train with the combined objective $\mathcal L = \mathcal L_{\mathrm{KR}} + \lambda \cdot \mathcal L_{\mathrm{CE}}$ for $\lambda \in \{0, 0.1, 1, 10, \infty\}$ where $\lambda = 0$ is pure kernel-residual and $\lambda = \infty$ is pure cross-entropy.
All other settings match the factorized sequence task experiment setup.
Results are reported for one seed per configuration.

\subsection{Seed Variability and Statistical Reporting}
\label{app:statistics}

All trained-model results in the revised sequence experiments use three
independent random seeds, $42$, $123$, and $2024$. Each seed controls
model initialization and stochastic training samples. Tables in
Appendix~\ref{app:corrected_sequence_results} report means across these
three runs.

Table~\ref{tab:seed_spread_main} reports the seed variability for the
main marginal-TV comparison. The improvements from Rao-Blackwellization
and the corrected residual weighting are larger than the observed
seed-to-seed variation across all six configurations.

\begin{table}[ht]
\centering
\caption{%
\textbf{Seed variability for the main corrected objective comparison.}
Exact one-step marginal TV, reported as mean $\pm$ standard deviation
over three seeds. Lower is better.
}
\label{tab:seed_spread_main}
\small
\begin{tabular}{lcccc}
\toprule
Data $(|\mathcal V|,D)$
& KR $(M{=}1)$
& KR-RB-uw
& PR
& KCE \\
\midrule
Bigram $(4,8)$
& $.0167\pm.0012$
& $.0024\pm.0004$
& $.0031\pm.0002$
& $.0027\pm.0009$ \\
Bigram $(8,16)$
& $.0233\pm.0031$
& $.0066\pm.0027$
& $.0053\pm.0023$
& $.0044\pm.0011$ \\
Bigram $(16,32)$
& $.0334\pm.0014$
& $.0053\pm.0002$
& $.0049\pm.0002$
& $.0038\pm.0001$ \\
Independent $(4,8)$
& $.0650\pm.0094$
& $.0031\pm.0005$
& $.0015\pm.0006$
& $.0010\pm.0002$ \\
Independent $(8,16)$
& $.0670\pm.0059$
& $.0074\pm.0004$
& $.0025\pm.0008$
& $.0023\pm.0003$ \\
Independent $(16,32)$
& $.0595\pm.0017$
& $.0137\pm.0005$
& $.0041\pm.0003$
& $.0033\pm.0003$ \\
\bottomrule
\end{tabular}
\end{table}

For the block experiment in
Section~\ref{sec:factorization_floor}, the reported values are also
three-seed means. At $|\mathcal V|=4$, $D=8$, and $\rho=0.9$,
Block-2 achieves total KL
$3.413\pm0.076$, within-block adjacent-pair TV
$0.0606\pm0.0013$, and across-block TV
$0.6751\pm0.0008$.
The corresponding tokenwise KR-RB-uw model has total KL
$7.2628\pm0.00003$.
For $|\mathcal V|=8$, $D=16$, Block-2 achieves
$12.873\pm0.871$ total KL,
$0.0668\pm0.0095$ within-block TV, and
$0.7874\pm0.0011$ across-block TV.

The second-moment experiment in
Section~\ref{sec:rb_experiments} does not train a model and therefore
has no training-seed variability. Each measured marker averages
$20{,}000$ independently corrupted samples, while the corresponding
curve is the closed-form prediction from
Eq.~\eqref{eq:second_moment}. Measured values agree with the closed form
to within $1.5\%$ except for four extreme cells with
$t\geq0.99$ and $M\leq4$, where the Monte Carlo estimate is itself
high variance.

Exact finite-state kernel recovery in
Section~\ref{sec:exact_kernel_recovery} is evaluated directly against
the analytical transition kernel. Additional multi-seed design
ablations are reported in Appendix~\ref{app:stage1}.

\subsection{Hyperparameter Summary}
\label{app:hyperparameters}

Table~\ref{tab:hyperparams} lists all hyperparameters used in the paper. The same optimizer and learning rate schedule are used across all experiments. Model capacity is scaled with task complexity.

\begin{table}[h]
\centering
\caption{Hyperparameters for all experiments.}
\label{tab:hyperparams}
\small
\begin{tabular}{@{}lll@{}}
\toprule
\textbf{Hyperparameter} & \textbf{Exact Kernel Recovery} & \textbf{Factorized Sequences} \\
\midrule
Optimizer & AdamW & AdamW \\
Learning rate & $3 \times 10^{-4}$ & $3 \times 10^{-4}$ \\
Weight decay & $10^{-5}$ & $10^{-5}$ \\
LR schedule & Cosine to 1\% & Cosine to 1\% \\
Gradient clipping & Max norm 1.0 & Max norm 1.0 \\
Batch size & 256 & 256 (KR, KCE), 128 (PR) \\
\midrule
Architecture & 4 layer MLP & 3 layer Transformer \\
Hidden dim & 128 & 128 \\
Attention heads & n/a & 4 \\
Time embedding dim & 32 (sinusoidal) & 128 (sinusoidal) \\
Dropout & 0.0 & 0.0 \\
\midrule
$\alpha$ schedule & $1 - \exp(-c(t-r))$ & $(t-r)/(1-r)$ \\
$\alpha$ parameter $c$ & $2 \times \max|u_t|$ & n/a \\
Time gap $\epsilon$ & 0.01 & 0.02 \\
\midrule
Training iters (2 state) & 15{,}000 & \\
Training iters (3 state) & 20{,}000 & \\
Training iters (10 state) & 30{,}000 & \\
Training iters ($|\mathcal V| \cdot D \leq 64$) & & 12{,}000 \\
Training iters ($|\mathcal V| \cdot D \leq 256$) & & 15{,}000 \\
Training iters ($|\mathcal V| \cdot D > 256$) & & 20{,}000 \\
\midrule
Random seeds & 42, 123, 2024 & 42, 123, 2024 \\
Evaluation samples & 5{,}000 & 5{,}000 \\
\bottomrule
\end{tabular}
\end{table}

No hyperparameter tuning was performed across configurations. The learning rate, batch size, and model dimensions were set once on a single pilot run and held fixed for all experiments and seeds. The only value that changes between exact kernel recovery configurations is the $\alpha$ parameter $c$, which is set deterministically from the reference generator and requires no tuning.

\subsection{Software and Libraries}
\label{app:software}

All experiments use PyTorch 2.5.1 with CUDA 12.1.
Matrix exponentiation for computing ground-truth kernels uses \texttt{scipy.linalg.expm} (SciPy 1.11).
Optimization uses the built-in PyTorch AdamW implementation.
No external generative modeling libraries are used.

\subsection{Broader Impact}

This work is foundational research on generative modeling in discrete state spaces,
validated entirely on synthetic data. It does not introduce new capabilities for
generating text, images, or other content that could be misused. If the framework
is eventually scaled to real language generation, it would inherit the same risks
as existing discrete generative models, including the potential for generating
misleading text. We do not foresee any immediate negative societal consequences
from the theoretical and experimental contributions presented here.

\end{document}